\documentclass[preprint, 12pt]{elsarticle}

\usepackage{amssymb}

\usepackage{amsmath,amsfonts}
\usepackage{colortbl}
\usepackage[table]{xcolor}
\usepackage{algorithmic}
\usepackage{algorithm}
\usepackage{array}
\usepackage[caption=false,font=normalsize,labelfont=sf,textfont=sf]{subfig}
\usepackage{textcomp}
\usepackage{stfloats}
\usepackage{url}
\usepackage{booktabs}
\usepackage{adjustbox}

\usepackage{hyperref}\hypersetup{
	colorlinks,
	linkcolor = blue,
	citecolor = blue,
	urlcolor=blue}
\usepackage[capitalize,nameinlink]{cleveref}
\usepackage{verbatim}
\usepackage{graphicx}

\journal{arxiv}

\begin{document}

\begin{frontmatter}



\title{Globalization for Scalable Short-term Load Forecasting}


\author[label1]{Amirhossein Ahmadi} 
\author[label1]{Hamidreza Zareipour}
\author[label1]{Henry Leung}

\affiliation[label1]{organization={Department of Electrical and Software Engineering, University of Calgary},
            addressline={2500 University Dr NW}, 
            city={Calgary},
            postcode={T2N 1N4}, 
            state={Alberta},
            country={Canada}}

\begin{abstract}

Forecasting load in power transmission networks is essential across various hierarchical levels, from the system level down to individual points of delivery (PoD). While intuitive and locally accurate, traditional local forecasting models (LFMs) face significant limitations, particularly in handling generalizability, overfitting, data drift, and the cold start problem. These methods also struggle with scalability, becoming computationally expensive and less efficient as the network's size and data volume grow. In contrast, global forecasting models (GFMs) offer a new approach to enhance prediction generalizability, scalability, accuracy, and robustness through globalization and cross-learning. This paper investigates global load forecasting in the presence of data drifts, highlighting the impact of different modeling techniques and data heterogeneity. We explore feature-transforming and target-transforming models, demonstrating how globalization, data heterogeneity, and data drift affect each differently. In addition, we examine the role of globalization in peak load forecasting and its potential for hierarchical forecasting. To address data heterogeneity and the balance between globality and locality, we propose separate time series clustering (TSC) methods, introducing model-based TSC for feature-transforming models and new weighted instance-based TSC for target-transforming models. Through extensive experiments on a real-world dataset of Alberta's electricity load, we demonstrate that global target-transforming models consistently outperform their local counterparts, especially when enriched with global features and clustering techniques. In contrast, global feature-transforming models face challenges in balancing local and global dynamics, often requiring TSC to manage data heterogeneity effectively.

\end{abstract}

\begin{graphicalabstract}
\includegraphics[width=1\textwidth,trim={0.15cm 0.15cm 0cm 0cm},clip]{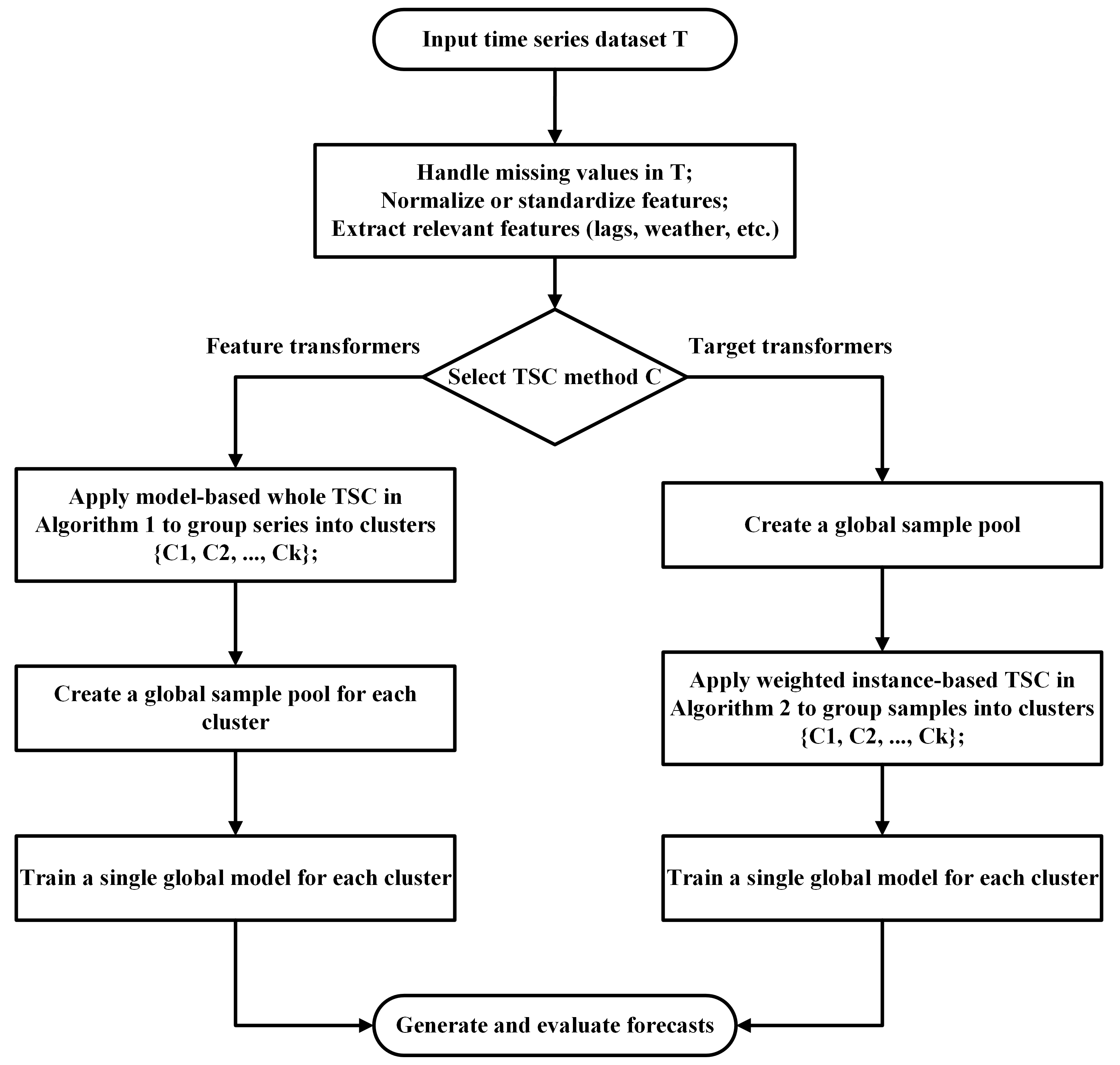}
\end{graphicalabstract}

\begin{highlights}
    \item A global modeling framework is proposed for the scalable forecasting of a large number of loads in a heterogeneous power transmission grid.
    \item Spatiotemporal sources of data heterogeneity besides data drift and demand response in Alberta's real-world electricity load are investigated through deep data analytics. 
    \item Different modeling techniques are analyzed, demonstrating how globalization, data heterogeneity, and data drift affect each differently.
    \item Two distinct TSC approaches are introduced, tailored for feature-transforming and target-transforming models.
    \item The impact of globalization on peak load and hierarchical forecasting is assessed, showing its potential for accurate zero-shot forecasting across multiple levels of the transmission network.
\end{highlights}

\begin{keyword}
Load forecasting \sep globalization \sep scalability \sep time series
clustering \sep weighted instant clustering \sep model-based clustering


\end{keyword}

\end{frontmatter}

\section{Introduction}

Power transmission networks are structured hierarchically, highlighting the need to forecast load not only at the system level but also at primary and secondary substations, area levels, and each point of delivery (PoD) \cite{ref1}. This requires flexible prediction techniques that can be applied to many measuring points with diverse characteristics while ensuring accurate and timely predictions \cite{ref2}. Traditionally, local forecasting has been used in a univariate way to address this challenge. This approach treats each time series as originating from a separate independent data-generating process \cite{ref3}, allowing for customized predictions at each measuring point. This type of forecasting is intuitive to understand and implement, allowing each model to be independently adjusted to explicitly account for the local behavior of individual time series, but it comes with several limitations \cite{ref4}.

Scaling to large datasets with numerous time series presents a significant challenge in real-world applications. This is because it requires ongoing efforts to retrain and manage many models in production, with each target variable necessitating its separate model \cite{ref88}. As the number of measuring points and the volume of data increase, local forecasting models (LFM) become computationally expensive and less efficient, often leading to deteriorated performance and slower response times \cite{ref7}.

The LFM also struggles with the forecasting of short time series, often experiencing overfitting due to the limited sample size \cite{ref5}. It also suffers from the cold start problem, which requires a substantial amount of historical data for each time series to estimate the model parameters reliably \cite{ref6}. Furthermore, it cannot predict new targets with few or even no observations \cite{ref7}, such as the energy consumption of a new PoD in a particular substation. It is also challenging for the local approach to quantify uncertainty and provide reliable prediction intervals \cite{ref8}. Finally, LFMs exhibit a deficiency in generalizability, failing to capture commonalities and dependencies among related time series, including cross-sectional or hierarchical relationships. This limitation is crucial to reduce uncertainty and improve the accuracy, generalizability, and robustness of forecasts \cite{ref9}.

Global forecasting is another forecasting paradigm that assumes the input time series are somewhat related to each other \cite{ref10}. In contrast to time series foundation models, which exploit relatedness within a single modality across multiple domains, global forecasting constrains this relatedness to a single domain \cite{ref11}. It can be implemented in two ways: data-centralized and data-distributed \cite{ref30}. The data-centralized approach aggregates all data into a unified global data pool at a central location, enabling the training of a single global model that captures shared patterns across all time series \cite{ref188}. However, it relies on data sharing, which can be restricted due to privacy or logistical challenges. In contrast, the data-distributed approach, often using frameworks like federated learning, trains local models independently on distributed data sources \cite{ref199}. Instead of sharing the data, only model parameters are aggregated into a global parameter pool, ensuring data privacy while collaboratively constructing a high-performance global model \cite{ref200}.

Unlike multivariate forecasting, global forecasting does not imply any causality or interdependence among the pooled time series in relation to the forecasts \cite{ref11}. Instead, it indicates that the parameters are estimated collectively for all available time series \cite{ref10}. Hence, multiple time series are used to increase the width of the dataset, keeping the length the same, and train a single global predictive model \cite{ref12}. By doing so, global forecasting models (GFMs) have a larger training set, and they can leverage shared structures across the targets to learn complex relations, ultimately leading to better predictions \cite{ref13}. By learning across many time series (cross-learning), GFMs can achieve a better generalization ability \cite{ref14}. The extra available time series also represents more training data, improving the overall forecasting performance (accuracy and robustness) \cite{ref15}.

Furthermore, GFMs can predict time series with little or no data, effectively addressing the cold start problem LFMs face \cite{ref6}. They have an implicit sense of similarity between different time series, enabling them to use similar patterns they have seen in time series with a rich history to develop a forecast on the new time series \cite{ref16}. By enforcing common parameters like seasonality across multiple time series, each with some noise, cross-learning acts like a regularizer, making GFMs more robust \cite{ref17}. Finally, GFMs provide scalability that significantly streamlines the training and deployment process by leveraging a single global model trained across multiple time series \cite{ref18}.

While GFMs yield reasonably good results, a notable drawback arises from their inductive bias, which assumes that the input time series are inherently related \cite{ref19}. This relatedness is typically defined as having identical or correlated data-generating processes \cite{ref3, ref20}. However, this assumption overlooks potential spatiotemporal data heterogeneity within a dataset and even within individual time series \cite{ref21}. Thus, if there are distinct groups of measuring points, the load series within each group may positively contribute to accurate predictions within their group but may negatively affect predictions in the other group due to spatially heterogeneous data-generating processes \cite{ref22}. Similarly, within an individual time series, traditional consumers might transition to modern prosumers equipped with solar PV panels, energy storage systems, or electric vehicles, leading to temporal heterogeneity through data drift or concept drift \cite{ref22}. External events such as wildfires, extreme weather, or pandemics like COVID-19 can also introduce distribution shifts. Hence, a single GFM might struggle to accurately capture the behavior of diverse time series (locality) and may require substantial memory and model complexity \cite{ref23, ref24}. Therefore, understanding data relatedness, data heterogeneity, and the interplay between globality and locality is crucial in the GFM context.

To address heterogeneity and strike a balance between globality and locality, a promising solution is to implement time series clustering (TSC) and fit a GFM for each distinct cluster \cite{ref233}. One approach is feature-based TSC as proposed in \cite{ref15}, where various features such as statistical properties and patterns are extracted from time series data. This method allows the model to focus on groups of time series with shared features, enhancing the model’s learning and prediction accuracy \cite{ref25}. However, feature-based clustering does not guarantee optimal partitions regarding relatedness or overall prediction accuracy \cite{ref25}. An alternative is using forecasting accuracy as a similarity measure through a feedback mechanism \cite{ref25}. This closed-loop whole TSC structure iteratively refines clusters based on prediction performance, ensuring tailored models for each cluster and improved forecasting outcomes within their specific contexts \cite{ref25}. Nonetheless, this approach overlooks the critical role of the forecasting model’s mechanism in clustering and globalization effectiveness.

We propose two distinct TSC approaches designed to address the challenges associated with feature-transforming and target-transforming forecasting models. Feature-transforming algorithms, such as linear regression and neural networks, learn a mathematical function that combines input features to predict target values, enabling extrapolation beyond the training set and adapting to data drift \cite{ref24}. In contrast, target-transforming algorithms, such as decision trees and nearest neighbors, utilize input features to group and average target values from the training set, resulting in predictions confined to the observed range \cite{ref24}. These fundamental differences necessitate tailored TSC methods to effectively handle heterogeneity and enable globalization.

For feature-transforming models, we propose a model-based clustering approach. Local models are trained for each time series, and their coefficients—such as the weights from linear regression—are used to cluster series based on process-driven similarities. This approach leverages the learned relationships between features and targets to group series with shared underlying patterns. For target-transforming models, we develop a weighted instance-based clustering method. Here, global model coefficients are utilized for weighting feature importance, enabling a weighted Euclidean distance metric that captures series similarity more effectively by accounting for the varying significance of features.

These proposed clustering methods allow us to analyze how globalization, data heterogeneity, and data drift impact the performance of different forecasting models. It aims a balance between locality, by preserving the unique characteristics of individual series, and globality, by leveraging shared patterns across the dataset. By addressing the fundamental differences in how feature-transforming and target-transforming models operate, our framework provides a scalable solution for load forecasting in heterogeneous transmission networks. This work goes beyond existing studies such as \cite{ref20} and \cite{ref30}, which focus on determining conditions for the effectiveness of global versus local models. Instead, we explore the influence of forecasting models' operating mechanisms on globalization performance.

To validate our approach, simulations demonstrate that GFMs maintain high accuracy levels comparable to local models while offering scalability benefits. Data analytics on real-world electricity load data from Alberta further uncovers spatiotemporal sources of data heterogeneity and drift, providing actionable insights to address challenges in dynamic power systems. Moreover, our TSC approaches for feature-transforming and target-transforming models enhance performance in critical scenarios, such as peak load and zero-shot forecasting across multiple levels of the transmission network, underscoring the scalability and adaptability of the proposed framework.

In summary, this paper makes the following key contributions:
\begin{itemize}
    \item It provides a comprehensive spatiotemporal analysis of data heterogeneity and drift using real-world electricity load data from Alberta.
    \item It proposes a global modeling framework for scalable short-term load forecasting across heterogeneous transmission networks with residential, commercial, and industrial areas.
    \item It introduces two tailored time series clustering approaches: a model-based clustering for feature-transforming models and a new weighted instance-based clustering for target-transforming models.
    \item It evaluates the impact of data heterogeneity, data drift, and globalization across different modeling techniques.
    \item It demonstrates the effectiveness of the proposed framework for peak load forecasting and zero-shot generalization across hierarchical levels of the power grid.
\end{itemize}

This manuscript is organized into the following sections. The literature is reviewed in \cref{sec:LiteratureReview} while different modeling concepts are formulated in \cref{sec:ProblemStatement}. Following this, two clustered global models for short-term electricity load forecasting are presented in \cref{sec:ProposedMethod}. Through several experiments using a real-world dataset of Alberta electricity load data, we demonstrate the effectiveness of the proposed approaches in forecasting load and particularly peak load in \cref{sec:Simulations}. Finally, \cref{sec:Conclusions} concludes the paper and elaborates on future research directions.

\section{Literature Review}  \label{sec:LiteratureReview}

The findings of the literature \cite{ref27, ref28, ref29} indicate that a single global load forecasting model generally outperforms individual local load forecasting models. However, to further address data heterogeneity, two common strategies are often used \cite{ref30}. The first is data pooling or clustering, which segments the data into more homogeneous subdatasets to improve performance by reducing variability. The second strategy involves specialization (or personalization), where the global model is enhanced with local components tailored to specific subsets, allowing the model to better capture localized patterns and nuances.

Shi et al. \cite{ref31} introduced a deep recurrent neural network based on random clustering for household load forecasting, showcasing the benefits of clustering mechanisms. However, the random approach can suffer from accuracy loss due to data distribution discrepancies. Han et al. \cite{ref32} improved this by using k-means clustering to group customers by load similarities, reducing overfitting and enhancing prediction accuracy. Building on this, Zang et al. \cite{ref33} applied mutual information for pooling interconnected users, further refining load forecasting through increased data diversity. Similarly, Yang et al. \cite{ref34} used hierarchical clustering and Bayesian deep learning for probabilistic load forecasting of customer groups.

Despite these benefits, these methods struggle to address concept drift, as the pools are formed based on static customer characteristics. Hence, Yang and Youn \cite{ref35} proposed a distribution-aware temporal pooling framework that leverages data clustering to identify related time series. This framework dynamically assigns time series to different forecasting models, accounting for potential changes in data distribution over time. This approach underscores the importance of adaptive modeling in improving forecast precision. Similarly, Yang et al. \cite{ref36} presented a temporal data pooling method that constructs pools at the sample level using distribution inference, mitigating overfitting, adapting to concept drift, and performing well even with unseen customers.

Nevertheless, pooling-based methods often overlook the potential advantages of integrating both global and local modeling perspectives. Since each model is typically trained on a specific subdataset, these approaches may suffer from limited generalizability and reduced robustness when applied to diverse or unseen scenarios. To address this, a parallel line of research has explored hybrid and specialization-based strategies. For example, Fallah et al. \cite{ref37} introduced a meta-learning approach for constructing a meta-global model, which serves as a foundation for developing personalized models fine-tuned using local data via gradient descent. Building on this, Dinh et al. \cite{ref38} formulated personalized federated learning as a bi-level optimization problem, where the global model is optimized at the outer level, and each client’s personalized model is obtained at the inner level by minimizing a regularized local loss.

Arivazhagan et al. \cite{ref39} proposed a method that trains the base layers of a model with global data and fine-tunes the specialization layers using local data to better capture unique characteristics. Wang et al. \cite{ref40} proposed a transformer-based model capable of forecasting multiple types of loads simultaneously. The model employs a unified encoder to transform the input data into a latent space, capturing high-dimensional interaction features between multi-energy loads and their associated information using an effective global attention mechanism. Grabner et al. \cite{ref41} took a different approach by first constructing a global NBEATS model to learn from all data and then fine-tuning it for specific clusters, allowing for localized adaptation.

Although specialization enhances a model’s ability to capture local patterns, its effectiveness relies heavily on the quality and representativeness of local data. In some cases, it may even undermine the benefits of the global model, leading to inferior performance compared to a purely global approach \cite{ref30}. Qin et al. \cite{ref30} illustrated how specialization can enhance load forecasting by analyzing the generalization bounds of both personalized and global models. Through mathematical derivations, they demonstrated how fine-tuning the global model could yield specialization gains, using simple matrix calculations to quantify this improvement. Their experimental results showed that while specialization is typically advantageous, it does not always guarantee improved performance \cite{ref30}.

Despite these advancements, several important gaps remain unaddressed. First, existing studies often overlook how globalization, data heterogeneity, and concept drift uniquely affect different modeling techniques. Our findings show that globalization interacts differently with distinct model classes, necessitating tailored treatments for effective generalization.

Second, prior work has rarely investigated the spatiotemporal sources of data heterogeneity. Most studies focus primarily on residential consumption, limiting the generalizability of their methods. To overcome this limitation, our study incorporates residential, commercial, and industrial price-responsive customers, thereby introducing broader heterogeneity and enabling a more comprehensive evaluation framework.

Finally, an important gap in the literature is the lack of examination into how globalization impacts peak load and hierarchical forecasting. We address this by exploring its potential for accurate zero-shot forecasting across multiple levels of the transmission network. Peak load forecasting is vital for grid reliability and resource planning, while hierarchical forecasting enables coordinated decisions across system, regional, and local levels.

\section{Problem Statement} \label{sec:ProblemStatement}

Scalable electricity demand forecasting in the transmission network involves predicting multiple time series across various network levels and sites. Existing models, which typically approach each time series as an individual regression problem, become computationally prohibitive under these conditions due to the complexity of training, hyperparameter tuning, deploying, monitoring, and retraining multiple models \cite{ref42}. This underscores the need for new load forecasting models that can: (i) scale efficiently with the increasing number of time series, (ii) generalize effectively across time series with diverse statistical properties, (iii) tackle heterogeneity and data drift, and (iv) deliver robust yet reliable load forecasts for extensive groups of time series data.

Assume $\mathcal{Y} = \{\textbf{y}_{i, 1:l_i}\}_{i=1}^n$ represent a collection of $n$ univariate time series, where each series $\textbf{y}_{i, 1:l_i} = [y_{i,1}, y_{i,2}, \dots, y_{i,l_i}]^T \in \mathbb{R}^{l_i}$ contains $l_i$ observations. Here, $ y_{i,t} \in \mathbb{R}$ is the observed value of the $i$-th time series at time $t$, and $l_i$ is the total length of the series. Let $h \in \mathbb{N}^+$ define the forecasting horizon, which specifies how many steps into the future need to be predicted. The task of load forecasting involves predicting the future values of the time series $\textbf{y}_{i, l_i+1:l_i+h} = [y_{i,l_i+1}, y_{i,l_i+2}, \dots, y_{i,l_i+h}] \in \mathbb{R}^h$.

Let $X_i = \{x_{i,1}, x_{i,2}, \dots, x_{i,m}\} \in \mathbb{R}^{m \times p}$ represent the feature matrix for $i$-th time series, where $m$ denotes the number of samples derived from a given time series, and $p$ indicates the number of features. Similarly, let $Y_i = \{y_{i,1}, y_{i,2}, \dots, y_{i,m}\}\in \mathbb{R}^{m \times h}$ denote the target matrix. To clarify, the index $i \in \{1, \dots, n\}$ is consistent across the notation and refers to the $i$-th time series in the collection $\mathcal{Y}$. For each series $\textbf{y}_i$, a set of $m$ training samples is generated using a sliding window approach over the historical data. This results in a feature matrix $X_i \in \mathbb{R}^{m \times p}$ and a corresponding target matrix $Y_i \in \mathbb{R}^{m \times h}$, where each row corresponds to one forecasting instance. Note that the number of training samples $m$ can vary across time series depending on the original series length $l_i$ and the choice of window size, lag depth, and forecast horizon $h$.

\subsection{Local Forecasting}

In the LFM paradigm, each time series $i$ has its own individual model, with the parameters $\theta_i$ optimized independently for that specific series as:

\begin{equation}
    \hat{Y}_i = f_i(X_i; \theta_i) + \epsilon_i
\end{equation}

\noindent where $\hat{Y}_i$ represents the predictions, and $\epsilon_i$ denotes the residual errors. The parameters $\theta_i$ are learned by minimizing a loss function $\mathcal{L}$, typically designed to measure the discrepancy between the actual targets $\hat{Y}_i$ and their predictions $Y_i$, over all $m$ samples as:

\begin{equation} \label{equ:eq2}
\theta_i^* = \arg\min_{\theta_i} \sum_{j=1}^{m} \mathcal{L}(y_{i,j}, \hat{y}_{i,j})
\end{equation}

\noindent where $y_{i,j}$ and $\hat{y}_{i,j}$ denote the actual and predicted values for the $j$-th sample of time series $i$, respectively. This approach results in $n$ independent models, one for each of the $n$ time series, allowing each model to specialize in capturing the unique characteristics of its corresponding series. However, this can become computationally expensive for large-scale datasets.

\subsection{Global Forecasting}

In the global pooling approach, a single model is trained on the combined data from all time series, aggregated into a unified global data pool at a central location. Let $\mathbb{X} = [X_1; X_2; \dots; X_n]^T \in \mathbb{R}^{M \times p}$ represent the global feature matrix, where $M = n \times m$ represents the total number of samples pooled across all time series. Similarly, let the global target matrix be $\mathbb{Y} = [Y_1; Y_2; \dots; Y_n]^T \in \mathbb{R}^{M \times h}$. This data pooling approach enables the training of a global model where the parameters $\theta$ are optimized collectively:

\begin{equation} 
\hat{\mathbb{Y}} = f(\mathbb{X}; \theta) + \epsilon
\end{equation}

\begin{equation}  \label{equ:eq4}
    \theta^* = \arg\min_{\theta} \sum_{i=1}^n \sum_{j=1}^m \mathcal{L}(y_{i, j}, \hat{y}_{i,j})
\end{equation}

This paradigm reduces the total number of models from $n$ (in the local approach) to one, allowing the global model to learn shared temporal and cross-series relationships. While this approach can efficiently utilize the commonalities across time series, it assumes that all time series share similar dynamics and may struggle to capture unique patterns in individual heterogeneous series.

\subsection{Cluster-wise Global Forecasting}

Cluster-wise global forecasting addresses the trade-off between scalability and specificity by partitioning time series into clusters with shared patterns, enabling the model to better handle heterogeneity across groups. Hence, instead of training a separate model for each time series or a single model for all time series, the time series are grouped into $K$ clusters based on shared characteristics. A single global model is then trained for each cluster, allowing the model to leverage shared patterns within clusters while also accounting for heterogeneity across clusters.

Let $\mathcal{C}_k$ denote the $k$-th cluster of time series, where $k \in {1, 2, \ldots, K}$, and each cluster $\mathcal{C}_k$ contains $n_k$ time series. The feature and target matrices for the $k$-th cluster are $\mathbb{X}_k=[X_i]_{i \in \mathcal{C}_k} \in \mathbb{R}^{M_k \times p}$ and $\mathbb{Y}_k=[Y_i]_{i \in \mathcal{C}_k} \in \mathbb{R}^{M_k \times h}$, where $M_k = n_k \times m$ represents the total number of samples pooled across cluster $k$. Then, for each cluster $\mathcal{C}_k$, a global model with parameters $\theta_k$ is trained across all time series $i \in \mathcal{C}_k$:

\begin{equation}
\hat{\mathbb{Y}}_k = f_k(\mathbb{X}_k; \theta_k) + \mathbf{\epsilon}_k \end{equation}

\begin{equation}
    \theta^*_k = \arg\min_{\theta_k} \sum_{i=1}^{n_k} \sum_{j=1}^m \mathcal{L}(y_{i, j}, \hat{y}_{i,j})
\end{equation}

\noindent This approach reduces the total number of models to $1 < K < n$, with one model per cluster, striking a balance between the specificity of local models and the scalability of a single global model. By clustering the time series based on shared patterns, the method adapts to data heterogeneity, allowing for improved forecasting accuracy within each cluster. However, clustering and its performance depend on the modeling technique, which determines the granularity of heterogeneity.

\subsection{Model-aware Cluster-wise Global Forecasting}

\begin{figure}[!t]
  \centering
  \includegraphics[width=0.96\textwidth,trim={0.15cm 0.25cm 0cm 0cm},clip]{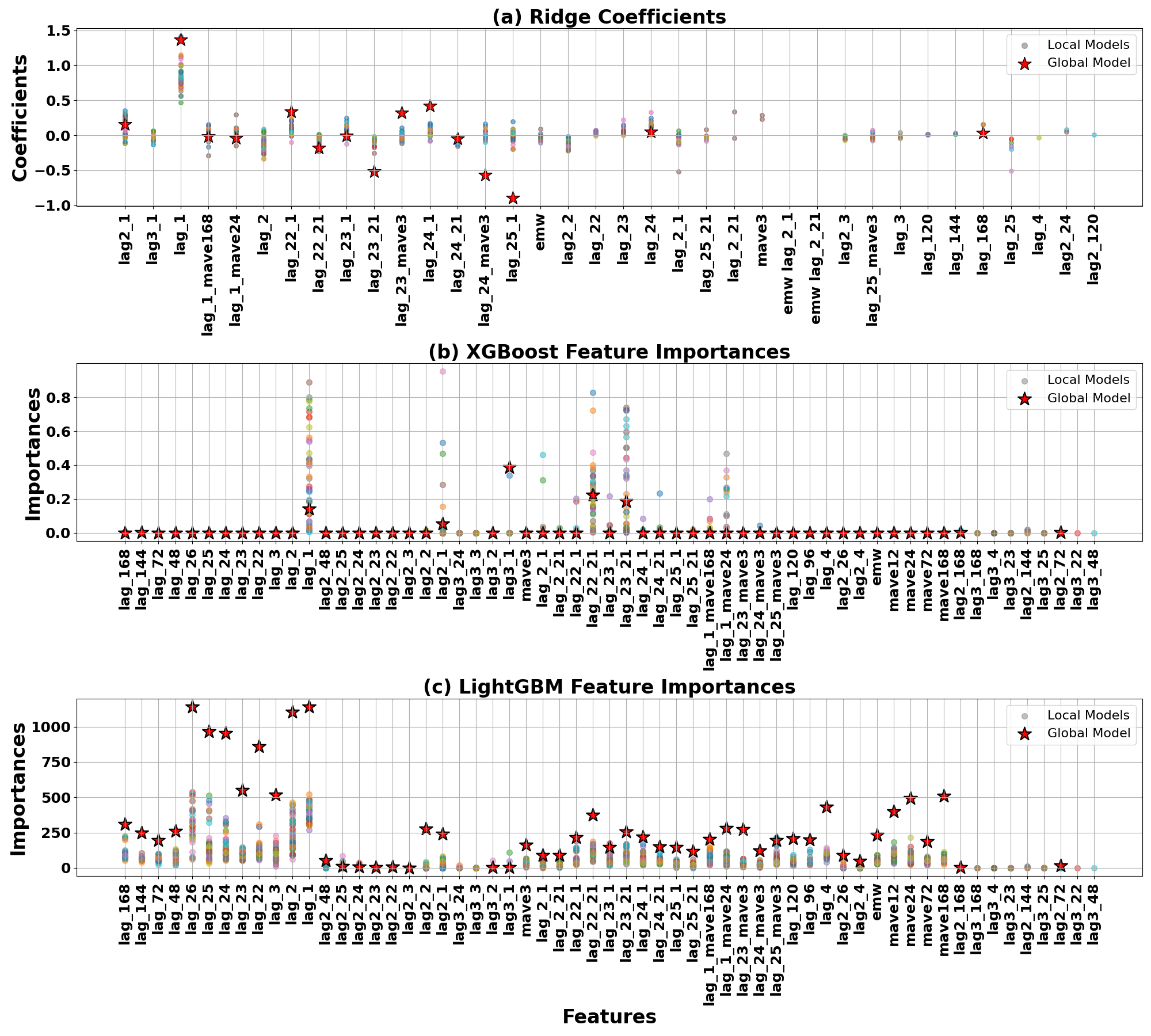}
  \caption{Model coefficients/importances comparison for LFMs and GFMs.}
  \label{fig:Fig2}
\end{figure}

In the context of globalization, applying a global model across multiple time series yields varying impacts depending on the algorithm's underlying nature. \cref{fig:Fig2} illustrates the differences between global and local models by comparing feature coefficients derived from Ridge regression and feature importances obtained from XGBoost and LightGBM. For feature-transforming algorithms, such as Ridge regression, \cref{fig:Fig2}a shows that globalization involves identifying common features across individual local models. By analyzing these localized patterns, a global linear model effectively aggregates and assigns weights to the most significant features shared across time series. This approach captures overarching trends or patterns that are consistent throughout the dataset, enabling the global model to generalize across diverse series while simplifying the modeling process.

However, fitting the same set of parameters to a diverse pool of time series often leads to a loss of granularity or locality in favor of globality, as the model sacrifices localized, specific behaviors to gain a broader, more general understanding \cite{ref20}. As a result, while the global feature-transforming model may generalize adequately across all series, it may lose accuracy on individual series where local behaviors are crucial \cite{ref17}. Hence, feature-transforming algorithms require more complexity through careful feature engineering and specific data partitioning techniques to be able to maintain accuracy. Thus, any TSC without considering the generating process of data can deteriorate the performance of globalization.

When it comes to target-transforming algorithms like LightGBM, \cref{fig:Fig2}c illustrates that globalization tends to work differently. LightGBM, being a decision-tree-based ensemble method, benefits from the increased diversity of pooled data that a global model provides. The model can learn from a larger pool of samples, allowing it to better generalize across different patterns and anomalies present in the time series. As LightGBM operates by creating multiple trees that consider various splits of the data, globalization enhances its ability to average over these diverse splits, leading to more robust predictions. The model may handle the heterogeneity in the data better by leveraging the ensemble's strength, where different trees in the ensemble might capture different aspects of the data's variability.

In this sense, globalization in a model like LightGBM does not just average feature importance like in linear models; instead, it builds a more comprehensive and nuanced understanding of the time series data by leveraging its ensemble nature. Each tree within the ensemble contributes to understanding different patterns, and globalization allows the model to combine these insights effectively. Hence, data partitioning for target transformers, though necessary to deal with data heterogeneity, reduces pool size and thus model performance. Hence, as with feature-transforming algorithms, careful TSC is necessary to improve pool quality in exchange for pool quantity and ensure that the model isn't overwhelmed by heterogeneity, which could lead to overfitting or underfitting depending on how the trees are constructed. This balance ensures that while the pool size is reduced through data partitioning, the relatedness and quality of the data will be improved, thereby enhancing the overall performance of target-transforming algorithms in a global model context.

\section{Proposed Method} \label{sec:ProposedMethod}

Generally, TSC algorithms can be classified as whole TSC, sequence, sample, and time point clustering based on using either a set of time series, subsequences of a single time series, or time points of a single time series, respectively \cite{ref43}. Given the distinction between target and feature transforming algorithms and to tackle data heterogeneity and provide a balance between locality and globality, we propose a model-based whole TSC approach for feature transforming forecasting models, specifically linear regression, and a weighted instance TSC for target transforming forecasting models, specifically decision trees.

\subsection{Model-based Whole TSC}

In the model-based TSC approach, the idea is to cluster time series with similar data-generating processes (DGPs), reflected by local coefficients, together. By grouping time series in this way, we aim to minimize the variance between the coefficients of the global model and those of the constituent local models within each cluster. This ensures that the global model can effectively capture the underlying patterns common to the time series in each cluster, leading to more accurate forecasting. The similarity in coefficients across time series within a cluster allows the global model to generalize well within that cluster, while still preserving the local nuances captured by individual models. This approach balances the need for locality in feature-transforming algorithms with the benefits of globalization, leading to improved model performance.

The overall process of the model-based TSC approach is summarized in \cref{alg:algorithm1} below, which provides a step-by-step breakdown of the proposed methodology. First, a local model is trained for each time series $i$ in the dataset, using the feature set $X_i$. Then, the local model parameters (coefficients) should be obtained as \cref{equ:eq2} to create a feature vector as

\begin{equation}
\text{Feature Vectors}: \theta_i = [\theta_{i,1}, \theta_{i,2}, \dots, \theta_{i,p}]
\end{equation}

\noindent where $p$ is the number of coefficients (or features) of the local models. Next, a clustering algorithm (here, K-means TSC) is applied to group time series based on their feature vectors as

\begin{equation}
\text{Cluster Assignment: } \mathcal{C}_k = \text{Cluster}(\theta_i)
\end{equation}

\noindent Finally, a global model is trained separately for each cluster using the time series within that cluster.

\begin{algorithm}[t]
\caption{Model-based Whole TSC}
    \label{alg:algorithm1}
    \begin{algorithmic}[1]
        \FOR{each time series $i$ in dataset}
            \STATE Train a local model: $\theta_i = \text{TrainModel}(X_i, Y_i)$
            \STATE Store coefficients $\theta_i$
        \ENDFOR
        \STATE Cluster time series based on $\theta_i$
        \FOR{each cluster $\mathcal{C}_k$}
            \STATE Train a global model $f_k$ using all time series in $\mathcal{C}_k$
        \ENDFOR
    \end{algorithmic}
\end{algorithm}

\subsection{Weighted Instance TSC}

In target-transforming algorithms like LightGBM, different features have varying levels of importance, as identified by the global model's coefficients. Thus, during clustering, not all features or variables will have the same level of importance. Hence, weighting in instance clustering is crucial as it serves as a form of feedback from the global model to accurately capture the relatedness between instances. By incorporating the global model's coefficients as weights in the clustering process, we can ensure that the most impactful features, according to the global model, play a larger role in determining the dissimilarity (distance) between instances. This feedback-driven approach allows the clustering algorithm to prioritize features that are globally significant while measuring distances, leading to more meaningful clusters that align with the true underlying patterns in the data.
 
To this end, a weighted instance-based TSC approach is designed for target-transforming forecasting models, such as LightGBM. The overall process of the proposed weighted instance-based TSC approach is summarized in \cref{alg:algorithm2} below, which provides a step-by-step breakdown of the proposed methodology. First, a global model is trained using the global dataset of all instances as \cref{equ:eq4}. Then, the global model coefficients that reflect the importance of each feature are extracted as

\begin{equation}
\theta = [\theta_1, \theta_2, \dots, \theta_p]
\end{equation}

\noindent Next, a weighted Euclidean distance metric is developed for instance clustering as

\begin{equation}
d_{i,j} = \sqrt{\sum_{r=1}^{p} \theta_r (x_{i,r} - x_{j,r})^2}
\end{equation}

\noindent Here, $d(i,j)$ is the distance between sample $i$ and $j$ weighted using global coefficients. Then, the time series are clustered based on the distance matrix $ \mathcal{D} = [d_{i,j}] \in \mathbb{R}^{M \times M}$ as

\begin{equation}
\mathcal{C}_k = \text{Cluster}(\mathcal{D})
\end{equation}

\noindent Finally, within each cluster, a global model is trained using the instances from that cluster.

\begin{algorithm}[t]
\caption{Weighted Instance TSC}
\label{alg:algorithm2}
    \begin{algorithmic}[1]
        \STATE Train a global model and obtain $\theta$
        \FOR{each sample $i$ in the global dataset}
            \STATE Compute weighted distance to other instances using $\theta$
        \ENDFOR
        \STATE Cluster instances based on distance matrix $\mathcal{D}$
        \FOR{each cluster $\mathcal{C}_k$}
            \STATE Train a global model $f_k$ using instances in $\mathcal{C}_k$
        \ENDFOR
    \end{algorithmic}
\end{algorithm}

\begin{figure}[!t]
  \centering
  \includegraphics[width=0.81\columnwidth,trim={0.15cm 0cm 0.1cm 0.15cm},clip]{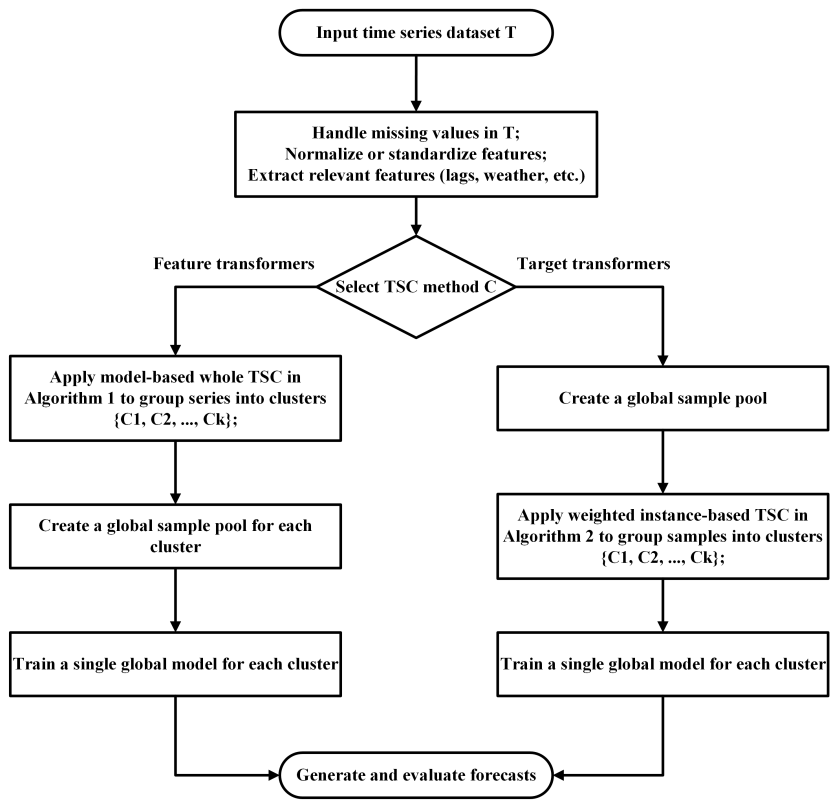}
  \caption{The proposed cluster-wise global forecasting flowchart}
  \label{fig:Fig1}
\end{figure}

\cref{fig:Fig1} illustrates the proposed methodology for cluster-wise global forecasting. Following preprocessing, the method diverges based on the selected TSC approach. For feature-transforming models, such as linear regression, the model-based whole TSC method is applied, and then separate global sample pools are created for each cluster. However, for target-transforming models, such as decision trees, a global sample pool is first generated to apply a weighted instance-based TSC method for grouping samples based on instance-level similarities, while assigning weights to prioritize important features. This approach ensures that feature-transforming models benefit from a focus on shared structural patterns across time series while target-transforming models leverage diverse instances within clusters for improved generalization. By integrating these tailored clustering methods, the proposed framework balances locality and globality, effectively addressing data heterogeneity and enhancing forecasting accuracy across different modeling paradigms.

\section{Simulation Results} \label{sec:Simulations}

The objective is to use the proposed method to develop cluster-wise global models for short-term electricity load forecasting. The real-world, open-access dataset utilized in this study comprises 42 time series derived from Alberta internal load (AIL), provided by the Alberta Electric System Operator (AESO) \cite{ref44}. It spans from January 2011 to October 2023 with an hourly resolution. Each time series corresponds to an individual area's hourly electricity load, and the dataset is structured with timestamps as the index and 42 area-level load values as columns.

\begin{figure}[!t]
  \centering
  \includegraphics[width=0.575\columnwidth,trim={0.35cm 0cm 0.1cm 0.5cm},clip]{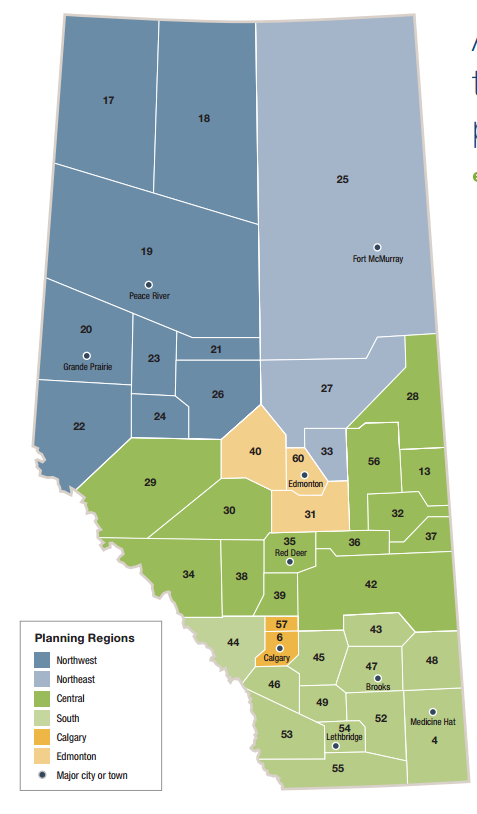}
  \caption{Alberta transmission planning areas \cite{ref45}.}
  \label{fig:Fig3}
\end{figure}

\cref{fig:Fig3} illustrates the Alberta transmission planning areas, which serve as the basis for this dataset. This figure highlights Alberta's geographical size and spatial diversity, characterized by considerable differences in climate, temperature, solar irradiation, humidity, and precipitation—all of which directly impact energy consumption and renewable generation patterns. The Alberta power network comprises 42 distinct areas grouped into six planning regions, each representing a unique zone for energy transmission. These regions collectively cover the entire province, enabling a comprehensive analysis of load forecasting across varied geographical and operational conditions. The load values are reported in megawatts (MW).

In this section, we present the numerical results for deterministic one-hour-ahead forecasting models. The data was split into training (2011–2021), validation (2022), and test (2023) sets before any normalization or sample construction to prevent data leakage strictly. A simple holdout validation strategy was employed, where a fixed year (2022) served as the validation window across all areas. This approach ensures consistent model comparison and avoids biases that could arise from varying error rates across time series with different seasonal structures or lengths. Moreover, it guarantees that every area contributes equally to the aggregate error metric, regardless of data availability.

Hyperparameter tuning was conducted using Optuna with hundreds of trials. For Ridge regression, we optimized the regularization parameter (alpha). For LightGBM, we tuned \texttt{n\_estimators}, \texttt{learning\_rate}, \texttt{max\_depth}, \texttt{num\_leaves}, and \texttt{boosting\_type}. For XGBoost, we tuned \texttt{n\_estimators}, \texttt{learning\_rate}, \texttt{max\_depth}, \texttt{max\_leaves}, and \texttt{booster}. After selecting the best hyperparameters, the models were retrained on the combined training and validation sets (2011–2022) and evaluated on the 2023 holdout test set. Across experiments, the default learning rates and booster settings, along with 200 estimators for local models and 1000 for global models, a maximum depth of 4, and 32 leaves, consistently yielded strong performance for both LightGBM and XGBoost. Ridge regression similarly performed well with a default regularization value of 1.

While this section focuses on deterministic one-step-ahead forecasting to highlight core contributions, the developed models can generate multi-step-ahead forecasts recursively. A detailed investigation into the performance of global models over extended forecasting horizons and under probabilistic settings is left for future work.

To evaluate the performance of our forecasts, we utilized normalized Mean Absolute Error (nMAE), Mean Squared Error (MSE), Mean Absolute Percentage Error (MAPE), and Forecast Bias (FB). nMAE and MSE measure absolute and squared errors, respectively, with MSE placing greater emphasis on larger errors, thereby reflecting overall accuracy while accounting for significant deviations \cite{ref18}. MAPE, a scale-free metric, evaluates percentage-based errors, making it suitable for comparisons across time series of varying scales \cite{ref18}. To complement these, FB provides a symmetric measure that highlights systematic over- or under-forecasting tendencies, offering a clear view of bias in the forecasts \cite{ref18}. This diverse selection of metrics ensures a comprehensive evaluation of accuracy, variability, and bias across a wide range of time series

\subsection{Data Analysis}

\begin{figure}[!t]
  \centering
  \includegraphics[width=1\textwidth,trim={0.15cm 0.25cm 0.15cm 1.5cm},clip]{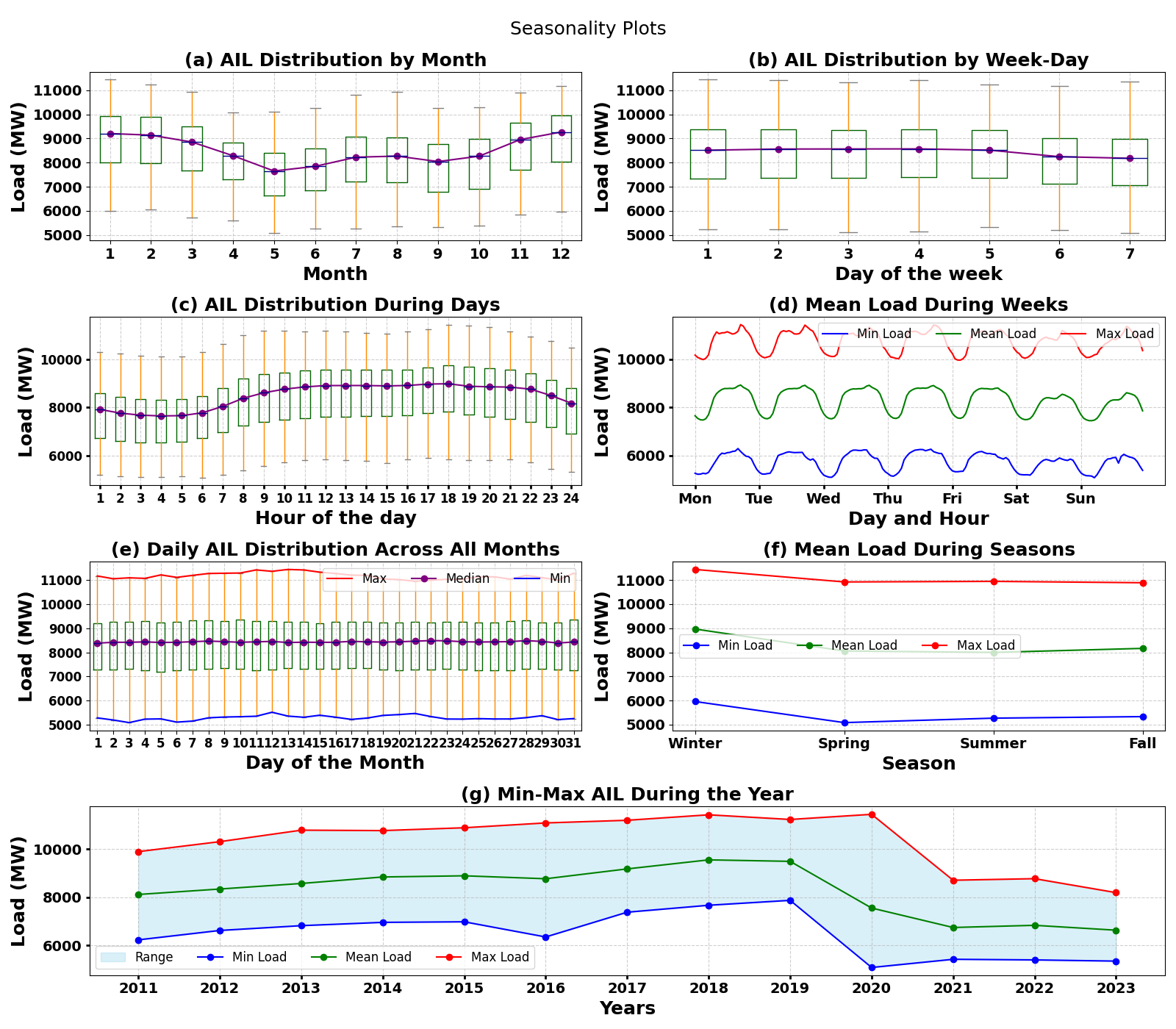}
  \caption{Load profile of Alberta internal load without considering transmission losses.}
  \label{fig:Fig4}
\end{figure}

Alberta's energy consumption is measured using the AIL with a hierarchical structure, divided into six planning regions and 42 areas \cite{ref46}. The sum of the regional or area-level loads represents the AIL without transmission losses \cite{ref47}. \cref{fig:Fig4} explores the seasonality and temporal variations in AIL across different dimensions such as months, days of the week, hours of the day, and seasons, providing a global picture of the energy consumption behavior in Alberta.

The seasonality analysis begins by examining the distribution of AIL across months, as shown in \cref{fig:Fig4}(a). The boxplot indicates that the AIL tends to be higher during winter (November to February), with January and December consistently exhibiting higher loads. This trend is reflective of increased heating demands during the colder months. The median AIL follows a clear seasonal pattern, peaking in January and dipping during summer, particularly in July and August. This pattern is corroborated by the corresponding plot, where the mean monthly load is superimposed, illustrating the cyclical nature of energy demand in Alberta.

The second panel in \cref{fig:Fig4}(b) shows the distribution of AIL by day of the week. This figure reveals a relatively consistent AIL throughout the week, with a slight reduction on weekends, particularly on Sundays. This decrease can be attributed to lower residential activity and reduced commercial operations over the weekend.

Further granularity is provided in the third panel \cref{fig:Fig4}(c), which explores AIL distribution by the hour of the day. Here, a clear diurnal pattern emerges, with AIL increasing steadily throughout the day, peaking in the early evening hours (around 6 PM to 8 PM), and declining during the night and early morning hours. This pattern aligns with typical residential and commercial energy usage, where demand rises as people return home from work and use household appliances.

The fourth panel in \cref{fig:Fig4}(d) examines the mean load distribution during weeks, capturing the interplay of daily and hourly trends across weekdays and weekends. This panel provides a detailed view of how AIL fluctuates within each day of the week, with higher loads observed on weekdays compared to weekends.

The fifth panel of \cref{fig:Fig4}(e) illustrates the daily distribution of AIL across all months, focusing on the load variation on different days of the month. A notable pattern emerges, with a peak in load consistently occurring around the 13th and 14th day of each month. This peak could be attributed to billing cycles or specific industrial activities that intensify around these dates, possibly due to production schedules or contractual obligations that align with mid-month deadlines.

The analysis also considers the variations in AIL across different seasons, as depicted in the sixth panel \cref{fig:Fig4}(f). The data demonstrates that winter consistently shows the highest load, followed by fall and spring, with summer exhibiting the lowest AIL. This seasonal variation is expected due to the high heating requirements during winter and moderate demands during transitional seasons.

In the final panel of \cref{fig:Fig4}(g), the annual variation in AIL is analyzed from 2011 to 2023. AIL gradually increased from 2011 to 2018, followed by a decline beginning in 2019. This trend coincides with the shift from coal to natural gas and the growing integration of renewables into the grid in Alberta \cite{ref46}. The year 2020 shows a significant decline in AIL, likely due to the impact of the COVID-19 pandemic, which resulted in reduced industrial activity and lower overall energy demand. However, the subsequent years show a recovery, although the levels remain below the pre-2019 peak.

\subsubsection{Data Heterogeneity}

Data heterogeneity in time series forecasting refers to the presence of diverse patterns, statistical behaviors, and structural characteristics either across different time series (spatial heterogeneity) or within different temporal segments of a single series (temporal heterogeneity) \cite{ref455}. This variability may arise from intrinsic differences in the underlying processes, such as region-specific consumption habits in power systems, or exogenous factors like weather, policy changes, behind-the-meter technology adoption, or infrastructure constraints. As highlighted in recent benchmarking studies, such heterogeneity poses significant challenges to the development of generalized forecasting models, as the assumption of a uniform data-generating process is often violated \cite{ref466}. Moreover, adversarial conditions like targeted data corruption can artificially introduce or amplify heterogeneity, further complicating model training and degrading predictive performance \cite{ref477}.

One significant source of data heterogeneity or dissimilarity in the data-generating process is the type of electricity consumption, which can be broadly categorized into residential, commercial, and industrial customers. These categories exhibit unique load patterns, particularly in their seasonal behavior, characterized by recurring variations over specific timeframes, such as hourly, daily, weekly, monthly, or annual cycles \cite{ref488}. Factors such as consumer behavior, climatic conditions, nighttime versus daytime differences, and weekday versus weekend activities contribute to these variations \cite{ref499}.

Haben et al. \cite{ref499} identified that analyzing average consumption during specific periods—overnight, breakfast, daytime, and evening—provides valuable insights into the characterization of residential customers. They developed seven features for customer clustering, which include the relative average load during these periods across the year, the annual mean relative standard deviation, a seasonal score, and a score indicating the difference between weekday and weekend consumption. Similarly, we developed seasonality index, total variation, night-to-day load ratio, and weekend-to-weekday load ratio tailored for the segmentation of industrial, commercial, and residential customers.

The seasonality index measures changes in the hour-of-day load profile, capturing the cyclical nature of energy consumption. Residential areas often exhibit significant seasonal and daily peaks due to factors like heating or cooling needs and varying occupancy patterns \cite{ref500}. In contrast, industrial areas display more consistent and stable load profiles due to the continuous nature of their production processes, which minimizes variability \cite{ref500}. Total variation quantifies overall load variability over time. Residential and commercial areas typically exhibit higher variation, reflecting fluctuations influenced by human activity and weather conditions \cite{ref500}. On the other hand, industrial areas tend to have lower variation, as their energy use is primarily dictated by stable, process-driven demands \cite{ref500}.

The night-to-day load ratio highlights differences in energy use between nighttime and daytime hours. Residential areas usually consume less energy during the night due to reduced activity, while industrial areas maintain steady energy usage throughout the day and night, often to ensure uninterrupted operations and meet production requirements \cite{ref500}. Similarly, the weekend-to-weekday load ratio measures variations in energy consumption between weekdays and weekends. Residential areas frequently show altered weekend patterns, with increased occupancy and leisure-related activities, whereas industrial facilities exhibit minimal changes, as they often operate 24 hours a day to avoid downtime and maintain efficiency \cite{ref500}.

\begin{figure}[!t]
  \centering
  \includegraphics[width=0.85\textwidth,trim={0.15cm 0.25cm 0.15cm 0cm},clip]{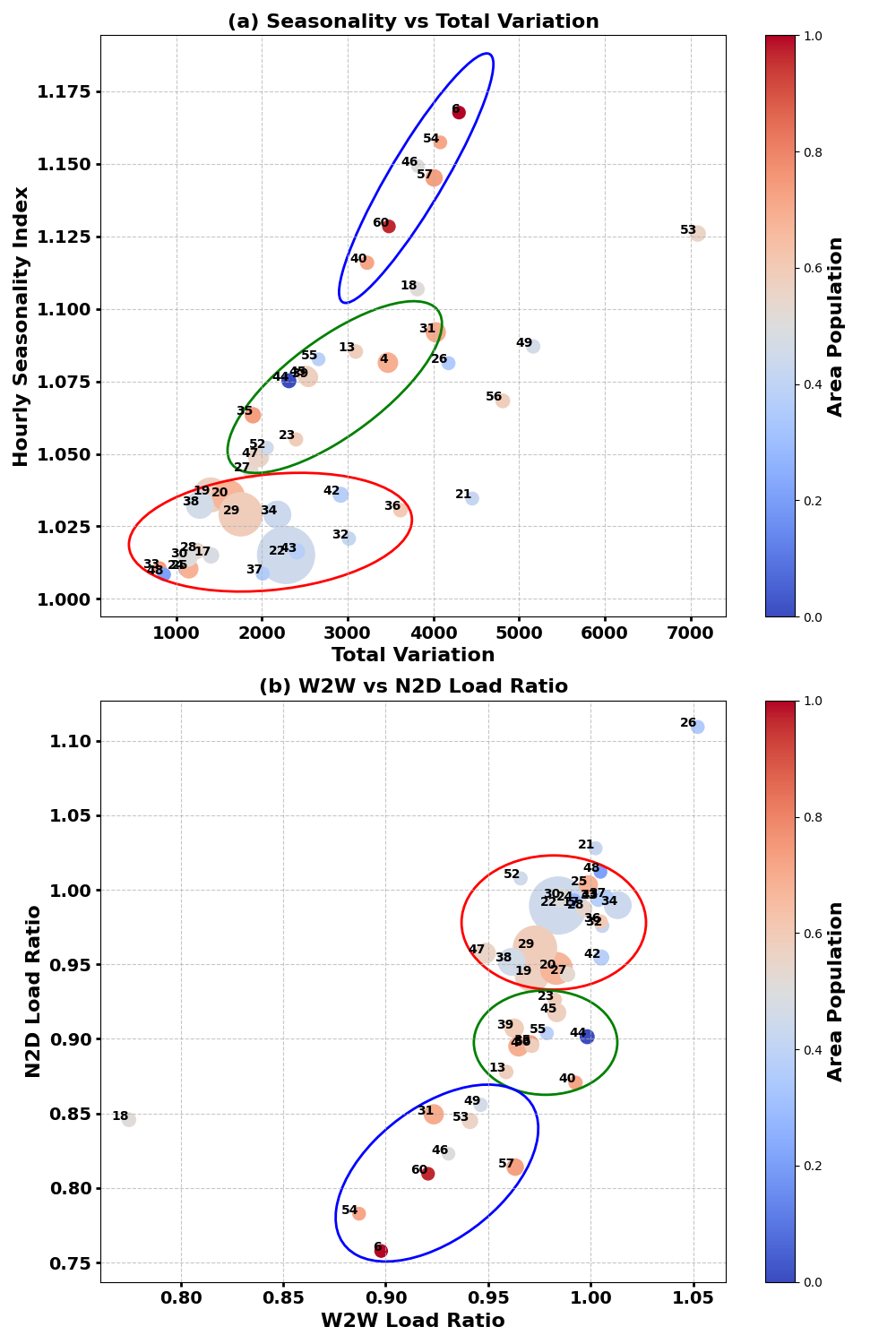}
  \caption{Data heterogeneity across Alberta's planning regions.}
  \label{fig:Fig5}
\end{figure}

\cref{fig:Fig5} demonstrates these distinctions visually, colored by population and sized by oil and gas production, showing how different areas align across these key metrics. Industrial areas like Cold Lake (area 28) and Grande Prairie (area 20) tend to cluster together with lower seasonality index and total variation and more uniform night-to-day and weekend-to-weekday load ratios, i.e., close to 1. In contrast, residential and commercial areas like Calgary (area 6) and Lethbridge (area 54) exhibit larger fluctuations in their consumption patterns across time, as evidenced by their higher seasonality index and more pronounced differences in load ratios.

The Northwest planning region, comprising areas such as 17, 18, 19, 20, 21, 22, 23, 24, and 26, is characterized by its status as a winter-peaking region, representing approximately 11\% of the total AIL \cite{ref46}. As shown in \cref{fig:Fig5}(a), the areas of this region exhibit a low total variation and seasonality index around, indicating limited fluctuations, besides concentrated night-to-day and weekend-to-weekday load ratios close to 1 as shown in \cref{fig:Fig5}(b), suggesting that these regions tend to have lower variability between weekdays and weekends and nighttime and daytimes. These characteristics reflect the region's industrial-heavy economy (oil and gas production, pulp and paper mills, etc.), contributing to relatively stable demand despite a smaller population (around 5\% of Alberta's total).

The Northeast planning region, which includes areas 25, 27, and 33, contributes approximately 30\% of the total AIL, even though it represents only 3\% of Alberta’s population \cite{ref46}. It is worth noting that Fort McMurray (area 25), as the region's largest city, has a population of nearly 77,000. The region's economy is largely driven by the oil sands industry, resulting in relatively low residential and commercial electricity loads. This region's consistent industrial operations resulted in a high and relatively stable energy demand, shown by low total variation and seasonality index and close-to-unit night-to-day and weekend-to-weekday load ratios in \cref{fig:Fig5}{a} and \cref{fig:Fig5}{b}, respectively.

The Edmonton planning region, comprising areas 31, 40, and 60, represents approximately 16\% of the total AIL and is home to about 34\% of Alberta’s population \cite{ref46}. This region is characterized by a mix of residential, commercial, and industrial loads, including oil refining, manufacturing, and pipelines. This balanced consumption profile leads to moderate to high variability and seasonality, influenced by both residential and industrial activities. Wabamun Lake area (area 40) is a major generating center with a large proportion of Alberta’s baseload coal-generating capacity.

The Central planning region is the largest of the planning regions, with 13 areas (13, 28, 29, 30, 32, 34, 35, 36, 37, 38, 39, 42, 56), representing 18\% of the total AIL \cite{ref46}. The economy is driven by a combination of oilfield production services, heavy oil production, manufacturing, farming, and pipeline activities. The region also includes significant oil sands and gas development in areas such as Cold Lake (area 28). Approximately 11\% of Alberta’s population resides in this region, which includes major population centers such as Red Deer (area 35) and Lloydminster (area 13). \cref{fig:Fig5} illustrates an almost industrial region with a low total variation and seasonality index and consistent load across days and weeks.

The Calgary planning region, consisting of areas 6 and 57, is a winter-peaking region that accounts for approximately 12\% of the total AIL \cite{ref46}. With 34\% of Alberta’s population residing in the region, electricity consumption is driven primarily by urban demand, including significant residential and commercial use, as well as low industrial activity. The region is characterized by higher daily and seasonal variability, especially due to residential heating and cooling needs. In Airdrie (area 57) and Calgary (area 6), nighttime and weekend loads are generally lower compared to daytime and weekdays, respectively, reflecting household activity patterns.

The South planning region, comprising areas 4, 43, 44, 45, 46, 47, 48, 49, 52, 53, 54, and 55, represents approximately 11\% of the total AIL \cite{ref46}. Approximately 12\% of Alberta’s population lives in this region, where the largest city in the region, Lethbridge (area 54), has a population nearing 107,000. Lethbridge shows high total variation and seasonality index with low night-to-day and weekend-to-weekday load ratios, reflecting its residential and agricultural energy demand. This region is characterized by its summer-peaking load profile, with electricity consumption driven by farming, agriculture, commercial (particularly meat and agri-food processing), and industrial activities such as pipelines and natural gas processing. The region’s energy demand also exhibits diverse load patterns due to a combination of industrial activities and residential consumption. The region is also home to the city of Medicine Hat (area 4) with a population nearing 67,000, which owns and operates its natural gas utility. It is known for its commercial activities, including cryptocurrency mining, which significantly impacts the load. The region also sees significant investment in renewable energy projects, particularly wind and solar power, further diversifying its energy demand.

This heterogeneity across the province underscores the importance of cluster-wise GFMs. Industrial-heavy regions exhibit less variability but higher overall consumption, while regions with more residential and commercial activity display significant daily and seasonal fluctuations. Addressing this heterogeneity is essential for scalable load forecasting and effective energy management across Alberta.

\subsubsection{Data Drift}

\begin{figure}[!t]
  \centering
  \includegraphics[width=1\textwidth,trim={0.25cm 0.25cm 0cm 0cm},clip]{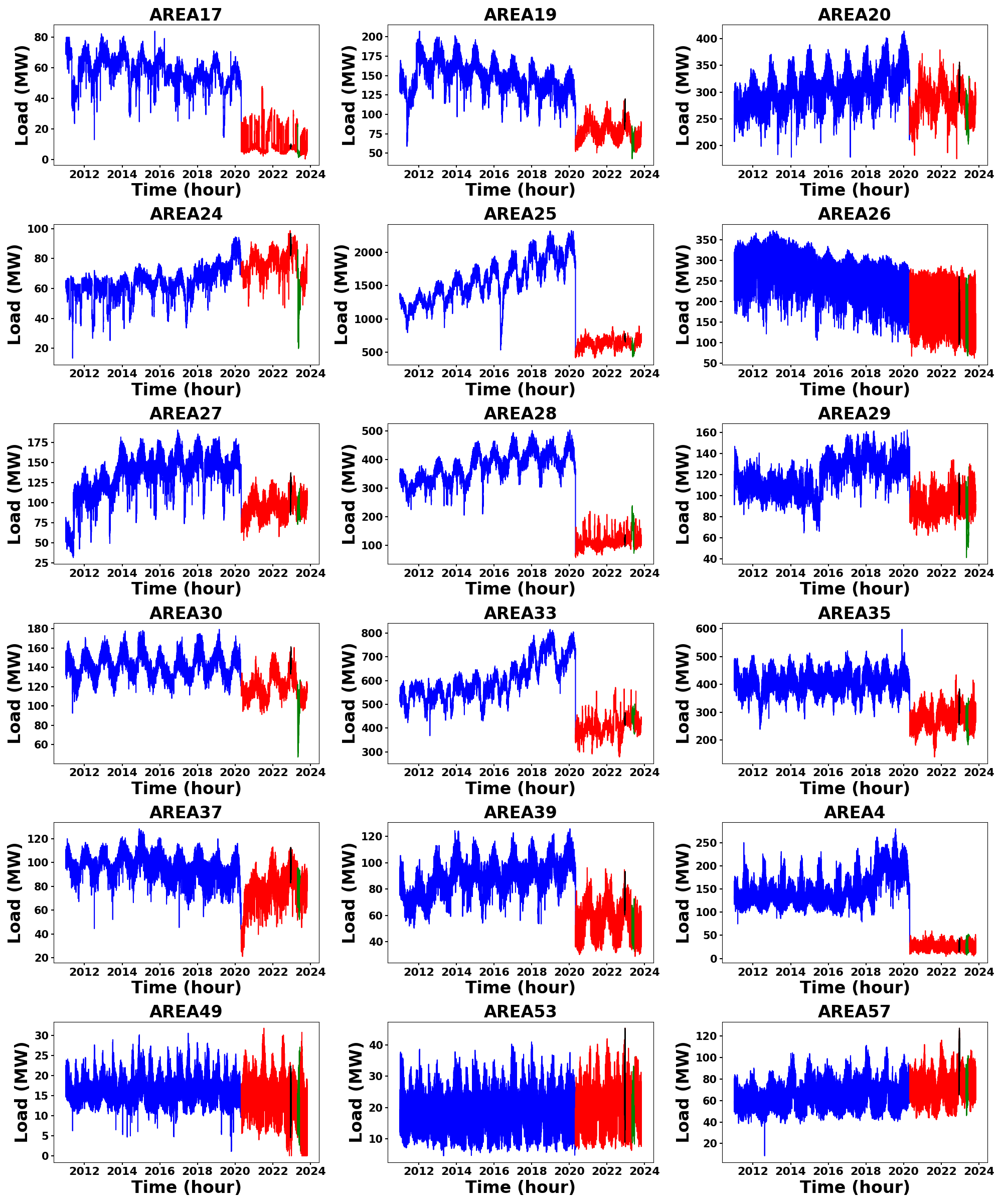}
  \caption{Data drift patterns across Alberta planning areas over time. The colored lines represent different periods of interest: blue indicates the pre-COVID-19 period, red corresponds to the COVID-19 period, green marks the wildfire period, and black denotes the extreme cold weather period.}
  \label{fig:Fig6}
\end{figure}

Recent events have underscored the profound impact of external factors on load patterns across Alberta's regions. \cref{fig:Fig6} introduces the changes in load patterns, highlighting a significant drop in May 2020 across several regions: Northwest (areas 17, 19, and 20), Northeast (areas 25, 27, and 33), Central (areas 28, 29, 33, 35, 37, and 39), and South (area 4). This decline, originating from the COVID-19 pandemic, is an example of sudden concept drift, where abrupt changes in human behavior, such as lockdowns and remote work policies, led to significant shifts in the load patterns.

In May 2023, the Northwest (areas 17, 19, and 24) and Central (areas 29, and 30) regions saw further load reductions of 5.2\% and 4.0\%, respectively, primarily due to wildfires \cite{ref47}. These natural events caused both temporary and long-term disruptions in electricity consumption. Wildfires represent another case of sudden concept drift, where abrupt disruptions affect load patterns and necessitate quick adaptation by utilities and consumers.

In contrast, predominantly residential and commercial, the Calgary planning region (area 57) experienced a slight 0.3\% decrease in system load \cite{ref47}. This modest reduction can be attributed to the increased installation of small-scale solar panels, such as rooftop solar, offsetting the underlying load growth from population increases \cite{ref47}. This scenario reflects an incremental drift, where gradual changes, such as adopting renewable energies, subtly alter the load profile over time.

The South planning region (area 49) experienced a 6.5\% decline in system load \cite{ref47}. However, this was mainly due to the addition of BTF generation rather than an actual loss of load, which was closer to 0.5\% and likely temperature-related \cite{ref47}. The region also saw a new winter peak in areas 53 and 57 due to extremely low temperatures and low wind generation, further illustrating the dynamic nature of load patterns and the importance of accounting for recurring drift in the load profile \cite{ref47}.

These data drifts underscore the critical need for model-specific TSC to account for the evolving nature of data distributions. Feature transformers can extrapolate target values beyond the range of the training data when given suitable input features. However, globalization deteriorates the performance of feature transformers during data drift, as the broader, more varied data may dilute the model's ability to accurately extrapolate based on specific, localized features. In contrast, target transformers are restricted to making predictions that remain within the bounds of the training set. Nevertheless, globalization improves the performance of target transformers during data drift by making the training set more representative of diverse conditions, thus allowing the model to better generalize across different scenarios. Thus, a model-specific approach is essential to optimize the benefits of globalization while mitigating its potential drawbacks, particularly in the context of data drift.

\subsubsection{Demand Response}

\begin{figure}[!t]
  \centering
  \includegraphics[width=1\textwidth,trim={0.15cm 0.25cm 0cm 0cm},clip]{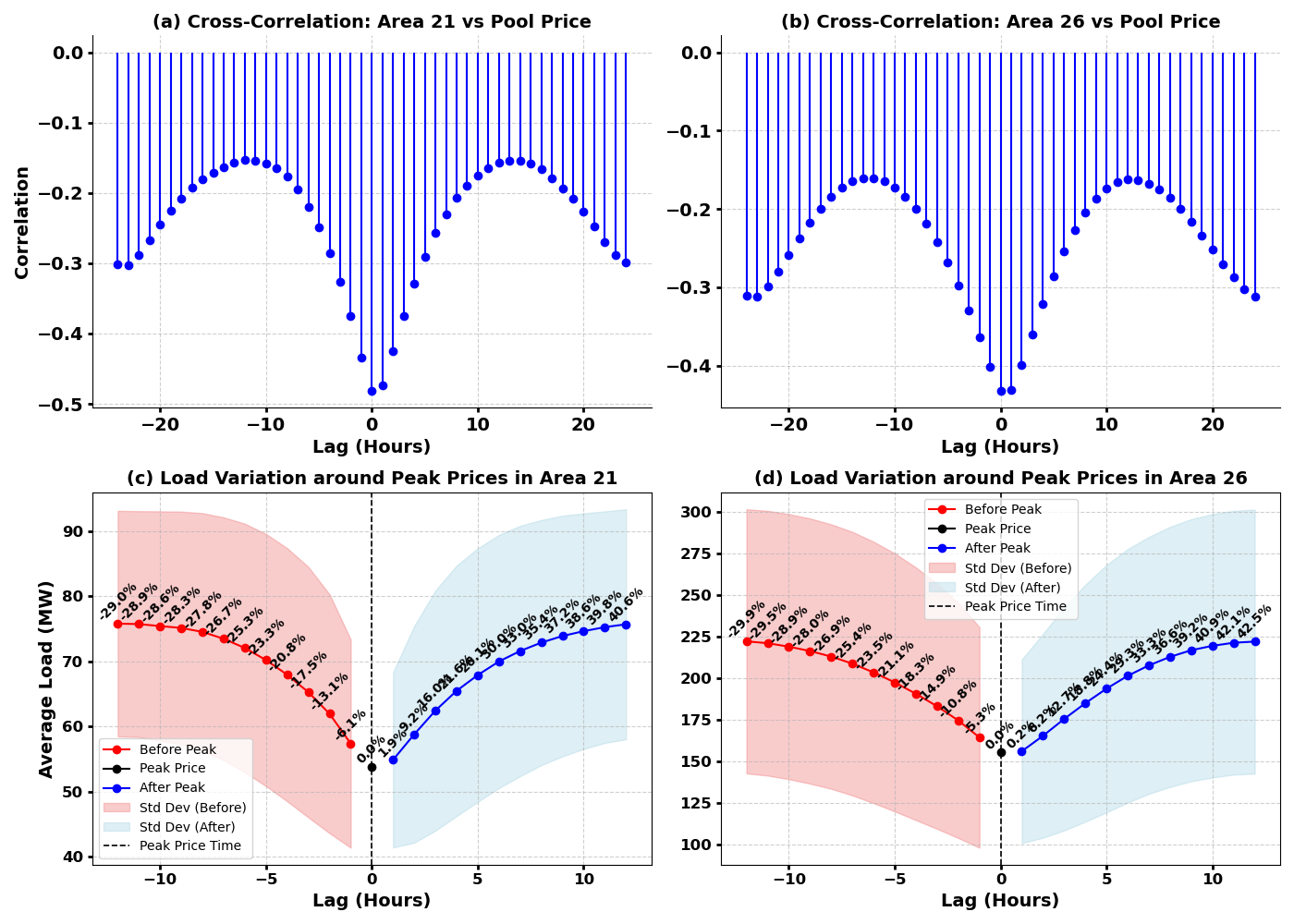}
  \caption{Cross-correlation and load variation analysis for Areas 21 and 26.}
  \label{fig:Fig7}
\end{figure}

The analysis of price responsiveness in Alberta provides valuable insights into how some areas adjust their electricity consumption in response to peak pool prices. \cref{fig:Fig7}(a) and \cref{fig:Fig7}(b) illustrate the cross-correlation between peak pool prices and load variations of areas 21 and 26 across different time lags, with a strong negative correlation at lag 0. This suggests that consumers in these areas promptly reduce their electricity usage when prices increase. The correlation structure also exhibits symmetry around lag -10 to +10 hours, indicating that price signals influence consumption decisions within a relatively short adjustment window. Such findings support the hypothesis that consumers in Areas 21 and 26 actively participate in price-responsive demand programs, demonstrating real-time load adjustments as a means of mitigating cost impacts.

Another analysis was conducted to further examine Areas 21 and 26's responsiveness to price signals by identifying high-price periods, defined as instances where the pool price exceeded the 95th percentile of its distribution. \cref{fig:Fig7}(c) and \cref{fig:Fig7}(d) illustrate the average electricity load variations in these areas relative to peak price events, with lags ranging from -12 hours before to +12 hours after the peak. The results reveal a consistent decline in load before peak price events, with Area 21 showing a maximum reduction of approximately 29\% and Area 26 experiencing a similar decline of nearly 30\% just before the price spike. The load gradually increases after the peak, indicating a rebound in consumption as prices stabilize. The shaded regions represent the standard deviation, highlighting the variability in load responses across different high-price events. At the moment of peak pricing (lag = 0), both areas exhibit a notable reduction in load, confirming the presence of demand-side elasticity. Area 21 records an overall decline of 6.15\% at peak time, while Area 26 sees a 5.28\% reduction. The post-peak recovery in load suggests that electricity consumption in these regions is highly sensitive to price changes, with consumers actively adjusting their usage patterns to mitigate high costs.

The combination of cross-correlation analysis and load variation studies provides compelling evidence of price responsiveness in Areas 21 and 26. The immediate reduction in consumption during high-price events, coupled with the correlation patterns observed at different time lags, confirms that consumers in these areas are sensitive to price fluctuations and adjust their electricity usage accordingly.

\subsection{Feature Engineering}

This section details the feature engineering process employed to enhance the complexity and generalization capability of the Global Forecasting Model (GFM). Theoretical evidence suggests that GFMs can perform comparably to Local Forecasting Models (LFMs) if sufficient attention is given to managing model complexity \cite{ref22}. Complexity in global models can be increased through the use of nonlinear or non-parametric models, representative feature construction, and data partitioning strategies \cite{ref22}. Since a GFM shares parameters across all series \cite{ref8}, comprehensive and expressive feature engineering is critical to ensure model adaptability across heterogeneous time series. Accordingly, we designed a rich set of features to capture nonlinearity, trend dynamics, seasonality, and external influences.

We employed a rolling window of 168 hours (7 days) and extracted a range of lag-based features. These include direct lag values at selected intervals (e.g., 1, 24, 48, ..., 168 hours), their squared and cubed values to model nonlinear temporal dependencies. To account for environmental dynamics, we incorporated weather-related features including temperature, dew point, wind speed, solar irradiance, cloud cover, and humidity. These features were aligned with the forecast target time and assumed to be accessible via hour-ahead forecasts, which is standard practice in operational forecasting. We further included higher-order temperature terms (e.g., $\text{temp}^2$, $\text{temp}^3$) and their moving averages over 3, 12, 24, 72, and 168-hour windows.

To capture complex relationships, we engineered interaction features involving lags, calendar variables, and temperature. Examples include interactions between lag and moving average (e.g., lag$_1$ $\times$ mave$_{168}$), between lags and their squared terms (e.g., lag$_2$ $\times$ lag$_1^2$), and between temperature and calendar cycles (e.g., temp $\times$ Hour, temp$^2$ $\times$ Month). These interactions help the model learn conditional dependencies across temporal and contextual dimensions.

We also introduced moving average features to smooth short-term fluctuations and reveal underlying trends in the load series. These were computed using fixed window sizes of 3, 12, 24, 72, and 168 hours, selected to capture sub-daily, daily, and weekly demand patterns. Additionally, an exponential moving average (EMA) with a span of 168 hours was used to emphasize recent observations while incorporating longer-term context.

Calendar features were used to encode periodicity and seasonality. Temporal variables such as year (offset from 2000), month, season, week of year, day of month, day of week, and hour were extracted and encoded using trigonometric functions (e.g., $\sin(2\pi \cdot \text{hour}/24)$, $\cos(2\pi \cdot \text{hour}/24)$) to preserve their cyclical nature. We also included a binary holiday indicator based on Canadian public holidays in Alberta, as well as a COVID-19 flag identifying whether a timestamp occurred during the pandemic period (May 1, 2020 – December 31, 2022).

External variables with predictive potential were included as exogenous inputs. These comprised the hourly Alberta electricity pool price (in \$/MWh), sourced from the AESO; daily West Texas Intermediate crude oil prices (in USD/barrel), resampled to hourly resolution via forward fill; and hourly Bitcoin (BTC/USD) prices from public APIs. Furthermore, we integrated outage-related indicators representing the hourly capacity outages (in MW) by generator type, i.e., coal, gas-fired, cogeneration, hydropower, wind, solar, and others, also sourced from AESO. The Load Shed Service Initiative (LSSI) variable, representing the amount of tripped load (in MW), was included as a continuous hourly feature sourced from AESO's outage reports.

To prevent data leakage, we ensured that all features used during model training and inference were aligned with or occurred before the prediction target time $t+1$. For features aligned at $t+1$, such as weather forecasts and exogenous market variables, we explicitly assume the availability of hour-ahead forecasts at prediction time. This is consistent with standard operational forecasting environments, where weather and market projections are routinely published and available at fine temporal resolutions. We have clarified this assumption in the manuscript to confirm that no post-target information was utilized during training, preserving the validity and reproducibility of our forecasting pipeline.

\cref{fig:Fig1} provides a comparison between the feature coefficients obtained from Ridge regression and feature importances derived from XGBoost and LightGBM models across different power planning areas in Alberta. \cref{fig:Fig1}(a)  illustrates the variability in Ridge coefficients for various features across all areas. Notably, features such as lag1, lag2, and exponential moving averages show significant fluctuations, reflecting their differing levels of influence in each area. The stars indicate the global model's coefficients, representing the global weights derived from all local models, highlighting shared important features across the dataset.

\cref{fig:Fig1}(b) displays the feature importances as determined by XGBoost across all areas, highlighting features that consistently exhibit high importance scores. Similarly, the bottom panel presents the feature importances derived from the LightGBM model, where these same features emerge as prominent in several areas, further reinforcing their significance in the modeling process. The stars in both panels represent the global model's feature importance, indicating the globally significant features that consistently hold relevance across the entire dataset. This alignment between the local and global models demonstrates the global model's ability to effectively aggregate and identify common influential features by averaging the importance scores from the local models.

However, \cref{fig:Fig1} discloses the fact that only a few features are utilized after globalization in the global model, highlighting the need for new features to better model commonalities between locals. To this end, polynomial feature generation is employed to capture non-linear relationships within the local time series data. Thus, interaction terms and polynomial features were generated up to the second degree. These features included interactions between lagged variables, moving averages, and exogenous factors such as temperature and calendar effects. 

As can be seen in \cref{fig:Fig1}, new cross-product terms consistently exhibited significant coefficients across multiple local models and were selected for inclusion in the final feature set. This approach ensured that the global model could leverage features that were both generalizable and relevant to specific time series patterns, thereby enhancing its overall performance.

\subsection{Local Load Forecasting}

\cref{Table1} compares the forecasting performance of LFMs across multiple metrics. Local Ridge emerges as the strongest performer in terms of overall consistency. Ridge demonstrates the smallest absolute bias, with a mean FB of 0.0004, suggesting that it is more balanced in its predictions compared to XGBoost and LightGBM. XGBoost shows some tendency towards over-prediction with a mean FB of 0.0013. LightGBM presents a similar pattern to XGBoost, with a slightly higher mean FB of 0.0013 and a maximum FB of 0.0220.

The nMAE values further support the superiority of Ridge in terms of predictive accuracy. With the lowest mean nMAE of 0.0214, Ridge shows smaller errors on average compared to XGBoost and LightGBM (both 0.0235), indicating better precision. Ridge also has a lower minimum nMAE (0.0087) compared to XGBoost (0.0101) and LightGBM (0.0100), highlighting its ability to provide more accurate forecasts in the best cases.

The MSE results also reveal that Ridge consistently delivers lower error magnitudes across the board. The mean MSE for Ridge (0.0008) is significantly smaller than that of XGBoost (0.0011) and LightGBM (0.0011), reinforcing Ridge’s superior error minimization. Also, Ridge achieves the lowest maximum MSE (0.0052), which indicates fewer large deviations from the actual values in comparison to XGBoost (0.0087) and LightGBM (0.0088).

While Ridge achieves the lowest minimum MAPE (1.38\%), it demonstrates less volatility in its predictions, as evidenced by its lower maximum MAPE (47.43\%) compared to both XGBoost (149.91\%) and LightGBM (148.00\%). Interestingly, Ridge’s mean MAPE (7.01\%) is more favorable than both XGBoost (12.85\%) and LightGBM (12.79\%), indicating that Ridge tends to produce lower percentage errors on average, though its worst-case errors are still notably large.

\begin{table}[t]
\centering
\caption{Normalized Performance Metrics for LFMs, GFMs, and Cluster-wise GFMs. The nMAE is calculated by dividing the MAE by the maximum value of the normalized ground truth for each area.}
\begin{adjustbox}{max width=\textwidth}
\begin{tabular}{llccccccccccc}
\toprule
\textbf{Metric} & \textbf{Model} & \multicolumn{3}{c}{\textbf{LFMs}} & \multicolumn{3}{c}{\textbf{GFMs}} & \multicolumn{3}{c}{\textbf{Model-based clustering GFMs}} \\
\cmidrule(lr){3-5} \cmidrule(lr){6-8} \cmidrule(lr){9-11}
 & & \textbf{Min} & \textbf{Mean} & \textbf{Max} & \textbf{Min} & \textbf{Mean} & \textbf{Max} & \textbf{Min} & \textbf{Mean} & \textbf{Max} \\
\midrule
\rowcolor{gray!20} \textbf{FB} & Ridge & \textbf{-0.0015} & \textbf{0.0005} & \textbf{0.0043} & \textbf{-0.0009} & \textbf{0.0002} & \textbf{0.0008} & \textbf{-0.0006} & \textbf{0.0003} & \textbf{0.0023} \\
                          & XGBoost & -0.0022 & 0.0013 & 0.0218 & -0.0024 & 0.0010 & 0.0212 & -0.0027 & 0.0011 & 0.0201 \\
                          & LightGBM & -0.0024 & 0.0013 & 0.0220 & -0.0025 & 0.0009 & 0.0179 & -0.0022 & 0.0011 & 0.0202 \\
\midrule
\rowcolor{gray!20} \textbf{nMAE (\%)} & Ridge & \textbf{0.8712} & \textbf{2.1417} & \textbf{6.5403} & \textbf{0.9318} & \textbf{2.1925} & 6.6524 & \textbf{0.9146} & \textbf{2.1509} & \textbf{6.4238} \\
                   & XGBoost           & 1.0115 & 2.3524 & 6.6742 & 1.0418 & 2.2837 & \textbf{6.3816} & 0.9483 & 2.2605 & 6.5652 \\
                   & LightGBM          & 1.0017 & 2.3538 & 6.6859 & 1.0452 & 2.2764 & 6.3967 & 0.9328 & 2.2685 & 6.4934 \\
\midrule
\rowcolor{gray!20} \textbf{MSE} & Ridge & 1.173e-05 & \textbf{0.0008} & \textbf{0.0052} & \textbf{1.038e-05} & \textbf{0.0008} & \textbf{0.0055} & \textbf{9.871e-06} & \textbf{0.0008} & \textbf{0.0052} \\
              & XGBoost & 1.074e-05 & 0.0011 & 0.0087 & 1.485e-05 & 0.0010 & 0.0092 & 1.557e-05 & 0.0010 & 0.0082 \\
              & LightGBM & \textbf{1.133e-05} & 0.0011 & 0.0088 & 1.439e-05 & 0.0009 & 0.0066 & 1.362e-05 & 0.0010 & 0.0086 \\
\midrule
\rowcolor{gray!20} \textbf{MAPE (\%)} & Ridge & \textbf{1.3832} & \textbf{7.0102} & \textbf{47.4379} & \textbf{1.5618} & \textbf{6.8582} & \textbf{35.7171} & \textbf{1.4982} & \textbf{6.7647} & \textbf{36.1005} \\
               & XGBoost           & 1.8123 & 12.8588 & 149.9138 & 1.6860 & 13.1145 & 253.532 & 1.8345 & 11.8836 & 180.257 \\
               & LightGBM          & 1.7654 & 12.7966 & 148.0044 & 1.6836 & 11.5732 & 196.498 & 1.6405 & 11.8409 & 166.165 \\
\bottomrule
\end{tabular} \label{Table1}
\end{adjustbox}
\end{table}

Auto-correlation experiment results indicated correlations across multiple lags, suggesting that the data exhibits a short-term linear dependence structure. Ridge regression, being a linear model with regularization, is particularly effective at handling such auto-correlations by controlling for overfitting while preserving linear relationships. In contrast, while XGBoost and LightGBM excel in capturing non-linear interactions, they may be less suited for data with dominant linear dependencies. This ability of Ridge to manage these linear dependencies through regularization likely explains its superior performance, as reflected in the metrics.

\begin{figure}[!t]
  \centering
  \includegraphics[width=1\textwidth,trim={0.15cm 0.25cm 0cm 0cm},clip]{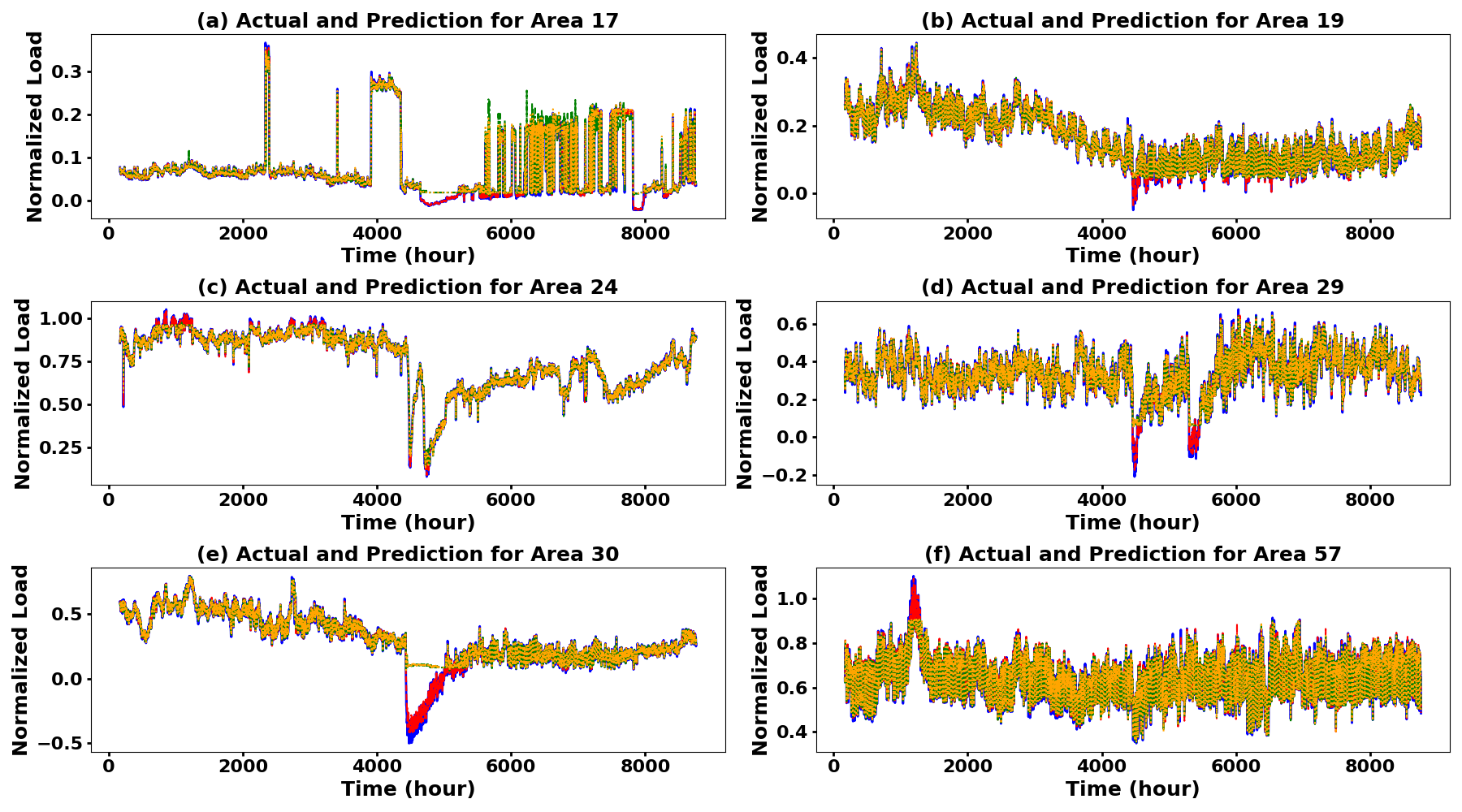}
\caption{Performance comparison of LFMs in the presence of data drifts. The actual electricity demand is shown in blue, while forecasts generated by Ridge, XGBoost, and LightGBM models are shown in red (dashed), green (dash-dot), and orange (dotted), respectively.}
  \label{fig:Fig8}
\end{figure}

Moreover, sudden changes in load patterns, as seen during the COVID-19 pandemic, extreme weather, or wildfires, caused the data to shift beyond the range of historical patterns. \cref{fig:Fig8} illustrates the extrapolation capability of LFMs during data drifts, highlighting the impact of sudden shifts in load patterns on forecasting accuracy. The figure shows that Ridge demonstrates better adaptability to these shifts compared to other models. This adaptability is due to Ridge's ability to extrapolate target values beyond the training set by leveraging the relationships in input features, as evidenced by its lower forecasting errors and bias. On the other hand, local target-transforming algorithms like XGBoost and LightGBM are confined to making predictions within the bounds of the training set. This limitation hinders their adaptability in the presence of significant data drifts, as they struggle to generalize to new or unseen conditions.

By analyzing the results across various areas, as shown in \cref{fig:Fig8}, particularly during peak hours, we notice a discrepancy between the actual observations and our predictions, indicating an offset. This suggests that the local models tend to replicate the most recent timestep (similar to a naïve forecast) rather than capturing the underlying patterns in the data. This tendency to mimic naïve forecasting is common in single-step-ahead models, as they are optimized solely for predicting the immediate next step \cite{ref18}. As a result, the model lacks the motivation to capture long-term dynamics, such as seasonal trends, and instead resorts to a behavior similar to naïve forecasting. In contrast, models designed to forecast over longer horizons are more likely to address this issue, as they are required to account for extended patterns in the data \cite{ref18}.

\subsection{Global Load Forecasting}

\cref{Table1} compares the forecasting performance of GFMs across multiple metrics. For the FB results of GFMs in \cref{Table1}, Ridge once again proves to be the most balanced model in terms of avoiding systematic over- or under-predictions, achieving the smallest mean FB of 0.0002. This slight positive bias suggests Ridge maintains a near-perfect balance in its predictions, with very minimal deviation from the actual values. The maximum FB for Ridge is also relatively small (0.0008), indicating that even in its worst cases, Ridge remains highly consistent. XGBoost and LightGBM, similarly, show lower levels of bias after globalization, with mean FB values of 0.0010 and 0.0009, respectively. These low values indicate reasonably balanced predictions overall for all global models.

The nMAE presents a slight increase for Ridge from 0.0214 to 0.0219, reflecting a marginal drop in precision, though it remains the best performer among the global models. Both XGBoost and LightGBM maintain very close nMAE values (0.0228 and 0.0227, respectively), with LightGBM slightly outperforming XGBoost in maximum nMAE. Globalization impacts MSE similarly, with a slight increase in this metric, though Ridge continues to achieve the lowest error. However, XGBoost and LightGBM both experience a decrease in MSE, with LightGBM reaching a lower maximum MSE than XGBoost, signifying fewer large deviations in some cases.

A key shift is observed in MAPE, particularly in the behavior of Ridge. Its MAPE drops slightly from 7.01\% in the local model to 6.86\% in the global model, showing improved consistency in percentage errors. Additionally, Ridge achieves the lowest maximum MAPE (35.72\%), confirming its robustness in avoiding extreme percentage errors. In contrast, XGBoost and LightGBM display significantly larger maximum MAPE values (253.53\% and 196.50\%, respectively), which indicates these models are more prone to occasional large forecasting errors in certain cases after globalization.

\begin{figure}[!t]
  \centering
  \includegraphics[width=1\textwidth,trim={0.25cm 0.25cm 0.25cm 0.25cm},clip]{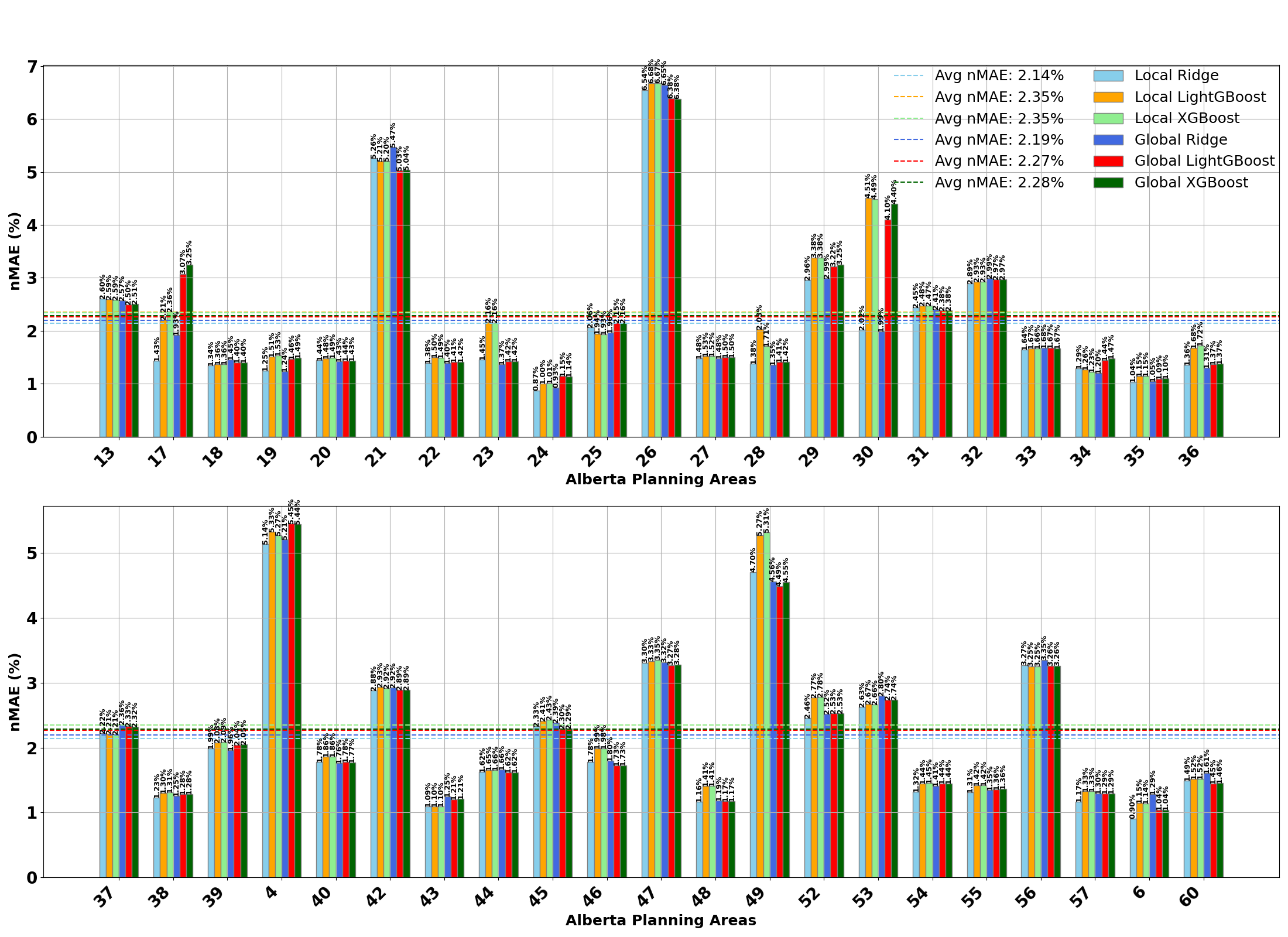}
  \caption{Performance comparison of LFMs and GFMs across different areas.}
  \label{fig:Fig9}
\end{figure}

Overall, \cref{fig:Fig9} demonstrates the impact of globalization by comparing the nMAE of local and global models across all 42 Alberta planning areas. On average, globalization improves the performance of tree-based models: both LightGBM and XGBoost show slight reductions in nMAE from 2.35\% (local) to 2.27\% and 2.28\% (global), respectively. This suggests that decision tree models benefit from increased data diversity and quantity when trained globally. In contrast, Ridge exhibits a slight increase in nMAE from 2.14\% (local) to 2.19\% (global), indicating that globalization introduces a minor loss of precision for feature transformers due to increased heterogeneity. Despite this, Ridge remains the best-performing model overall in terms of average nMAE, highlighting its robustness and strong generalization capabilities.

\begin{figure}[!t]
  \centering
  \includegraphics[width=1\textwidth,trim={0.15cm 0.25cm 0cm 0cm},clip]{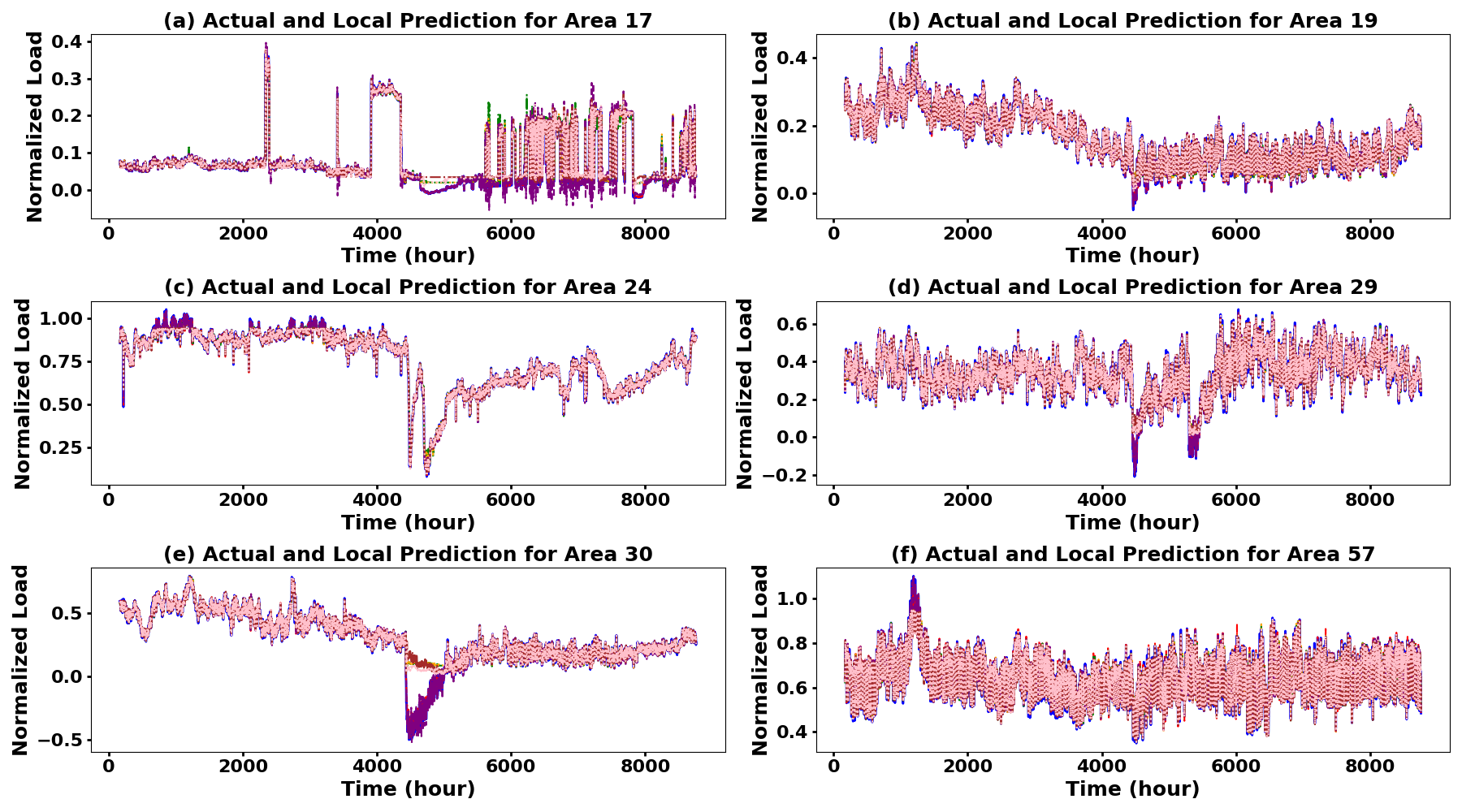}
\caption{Performance comparison of GFMs and LFMs in the presence of data drifts. The actual electricity demand is shown in blue. Forecasts from LFMs are represented as follows: Ridge in red (dashed), XGBoost in green (dash-dot), and LightGBM in orange (dotted). Forecasts from GFMs are shown as: Ridge in purple (dashed), XGBoost in brown (dash-dot), and LightGBM in pink (dotted).}
  \label{fig:Fig10}
\end{figure}

\cref{fig:Fig10} illustrates the performance of GFMs during data drifts, highlighting the impact on the extrapolation capability of models under globalization. The figure demonstrates that globalizing feature-transforming models like Ridge result in a noticeable degradation in their performance during data drifts. It means that the extrapolation capability of these models is degraded as they sacrifice local specificity for broader generalization.

In contrast, \cref{fig:Fig10} illustrates globalization benefits target-transforming models like XGBoost and LightGBM. These tree-based ensemble methods leverage the diversity of pooled data to generalize effectively across patterns and handle data drifts. By building multiple trees with varied data splits, these models capitalize on the increased sample diversity in a global context, resulting in improved extrapolation.

To further support these findings, \cref{Table22} compares the nMAE of global and local models separately in stable and drifting areas. In stable regions, both LightGBM and XGBoost exhibit clear improvements of +4.87\% and +4.74\%, respectively, while Ridge shows a negligible deterioration (-2.59\%). More notably, LightGBM continues to outperform its local version in drifting areas (+2.09\%), and XGBoost maintains a marginal improvement. Although Ridge performs well on average, its performance drops slightly in drifting areas (-2.15\%), highlighting the sensitivity of feature-transforming models to distributional shifts. These results emphasize the importance of selecting model types aligned with the degree of data drift and support the scalability and robustness of global target-transforming models in diverse environments.

Finally, when examining the global models' performance across different areas given in \cref{fig:Fig10}, we observe a similar misalignment between the actual observations and the forecasts, leading to a small offset. This issue indicates that GFMs, like LFMs, fall into the trap of replicating the most recent timestep, resembling a naïve forecast, instead of learning the true data patterns. Consequently, global models fail to capture long-term behaviors, such as seasonality, when trained for single-step-ahead forecasting and end up imitating the naïve forecast.

\begin{table}[t]
\centering
\caption{Comparing Normalized MAE (\%) for Drifting and Stable Areas}
\begin{adjustbox}{max width=\textwidth}
\begin{tabular}{llccccccccccc}
\toprule
\textbf{Metric} & \textbf{Model} & \multicolumn{3}{c}{\textbf{Stable Areas}} & \multicolumn{3}{c}{\textbf{Drifting Areas}} \\
\cmidrule(lr){3-5} \cmidrule(lr){6-8}
 & & \textbf{Local} & \textbf{Global} & \textbf{Change (\%)} & \textbf{ocal} & \textbf{Global} & \textbf{Change (\%)} \\
\midrule
\textbf{nMAE (\%)} & Ridge & 2.00 & 2.05 & -2.5917 & 2.33 & 2.38 & -2.1540 \\
            & XGBoost & 2.11 & 2.01 & 4.7362 & 2.66 & 2.64 & 0.4432 \\
           & LightGBM & 2.11 & 2.01 & 4.8692 & 2.67 & 2.61 & 2.0948 \\
\bottomrule
\end{tabular} \label{Table22}
\end{adjustbox}
\end{table}

\subsection{Cluster-wise Global Load Forecasting}
\subsubsection{Model-based Whole TSC}

\begin{figure}[!t]
  \centering
  \includegraphics[width=0.675\textwidth,trim={0.15cm 0.25cm 0cm 0cm},clip]{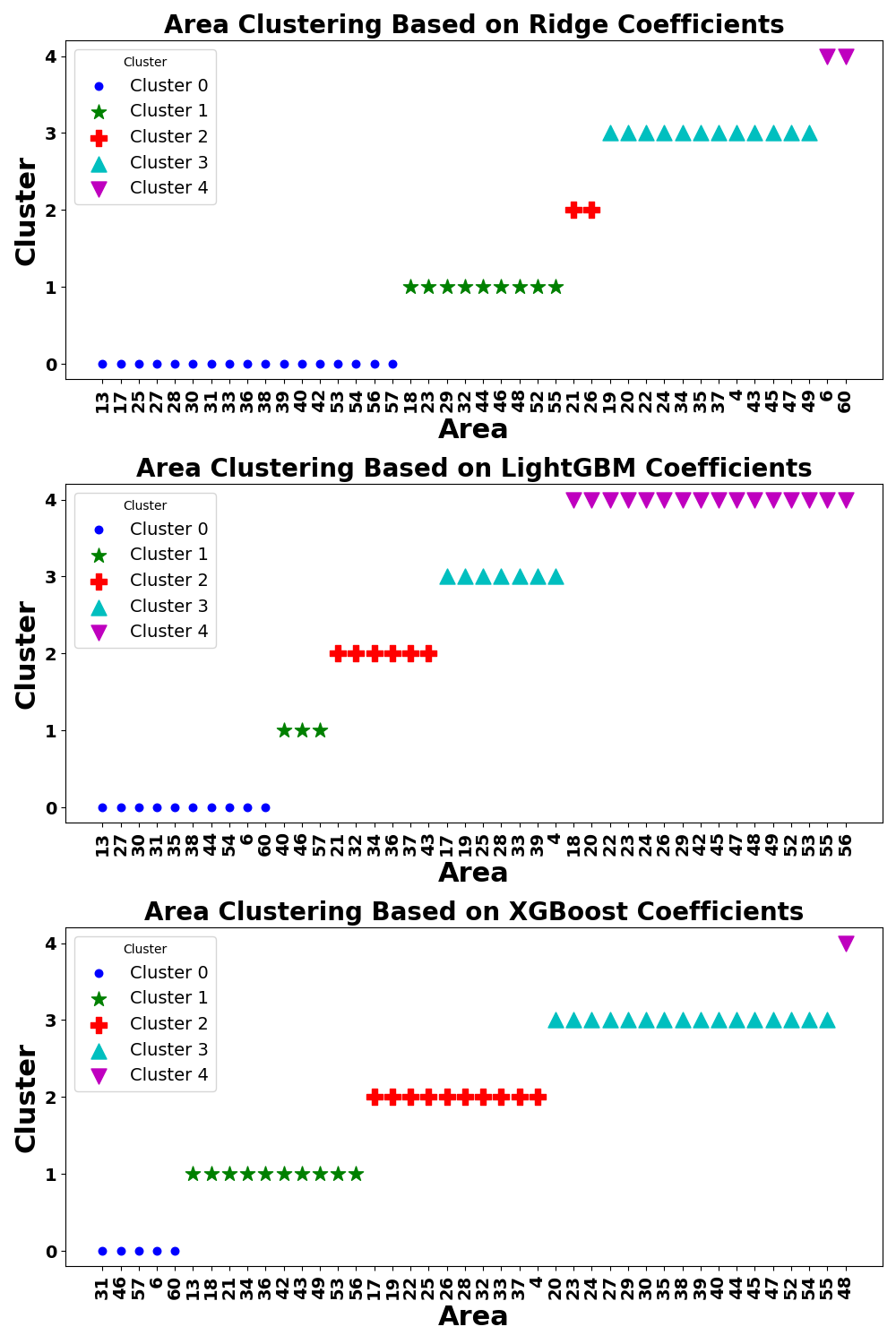}
  \caption{Area clustering based on model coefficients.}
  \label{fig:Fig11}
\end{figure}

\cref{fig:Fig11} illustrates the clustering of different areas based on the developed model-based whole TSC, i.e., model coefficients derived from Ridge, LightGBM, and XGBoost models. Each subplot represents the area clustering results for one of the models, showing how different areas are grouped into distinct clusters based on their feature coefficients/importance in the respective models. The scatter plots demonstrate the distribution of areas across various clusters, each represented by distinct markers and colors.

\cref{Table1} compares the forecasting performance of model-based whole TSC-wise GFMs across multiple metrics. The results indicate that model-based clustering had varying impacts across the different models. For LightGBM, the mean performance slightly deteriorated, as reflected in a marginal increase in error metrics like nMAE and MSE. In contrast, XGBoost showed a slight improvement in mean performance, with a minor reduction in some error metrics, suggesting that the clustering was somewhat effective for this model. However, the most noticeable improvement was observed with Ridge regression, where the mean performance metrics, such as nMAE, and MSE, all showed clear enhancements.

This pattern suggests that model-based whole clustering works particularly well for feature-transforming models like Ridge regression. The clustering enables Ridge to maintain its linear structure while adapting better to the localized patterns in each cluster, thereby improving its ability to generalize across the entire dataset. On the other hand, decision-tree-based models like LightGBM and XGBoost, which rely on pool size, benefit less from clustering. Whole TSC reduces the sample size within each cluster without considering the quality (homogeneity) of the samples, leading to a loss of the primary advantage of globalization—access to a larger, more diverse dataset. Thus, the target transformer’s mean performance deteriorated due to the decreased data diversity within each cluster.

\subsubsection{Weighted Instance TSC}

\cref{Table2} shows the performance of different GFMs under instance-based TSC approaches, where it brought improvements for target-transforming models. Unlike model-based TSC, which divides the data based on model parameters, instance-based TSC groups samples with similar characteristics, allowing the models to preserve diversity and improve the quality of the dataset. This helped XGBoost and LightGBM recover from the performance degradation observed in model-based TSC. XGBoost showed improvements in error metrics like nMAE and MSE. By retaining diverse instances within each cluster, the model was better able to capture non-linear patterns and generalize more effectively across various regions. Similarly, LightGBM benefited from the increased instance diversity, with its mean performance improving notably, particularly in error reduction. The clustering allowed LightGBM to continue leveraging its tree-based structure to capture complex relationships without losing the global context.

\begin{table}[t]
\centering
\caption{Normalized Performance Metrics for GFMs, and Instance-based Cluster-wise GFMs. The nMAE is calculated by dividing the MAE by the maximum value of the normalized ground truth for each area.}
\begin{adjustbox}{max width=\textwidth}
\begin{tabular}{llccccccccccc}
\toprule
\textbf{Metric} & \textbf{Model} & \multicolumn{3}{c}{\textbf{GFMs}} & \multicolumn{3}{c}{\textbf{Instance-based GFMs}} & \multicolumn{3}{c}{\textbf{Weighted Instance-based GFMs}} \\
\cmidrule(lr){3-5} \cmidrule(lr){6-8} \cmidrule(lr){9-11}
 & & \textbf{Min} & \textbf{Mean} & \textbf{Max} & \textbf{Min} & \textbf{Mean} & \textbf{Max} & \textbf{Min} & \textbf{Mean} & \textbf{Max} \\
\midrule
\rowcolor{gray!20} \textbf{FB} & Ridge & \textbf{-0.00096} & \textbf{0.00026} & \textbf{0.00080} & \textbf{-0.00126} & \textbf{-0.00010} & \textbf{0.00159} & \textbf{-0.00071} & \textbf{-0.00010} & \textbf{0.00166} \\
                             & XGBoost & -0.00247 & 0.00100 & 0.02118 & -0.00197 & 0.00244 & 0.02043 & -0.00579 & 0.00015 & 0.01816 \\
                            & LightGBM & -0.00250 & 0.00089 & 0.01792 & -0.01024 & -0.00403 & 0.01239 & -0.00514 & -0.00209 & 0.01452 \\
\midrule
\rowcolor{gray!20} \textbf{nMAE (\%)} & Ridge & \textbf{0.9318} & \textbf{2.1925} & 6.6524 & \textbf{0.9228} & \textbf{2.1856} & 6.5603 & \textbf{0.9318} & \textbf{2.1925} & 6.6524 \\
                                    & XGBoost & 1.0418 & 2.2837 & \textbf{6.3816} & 1.0902 & 2.2659 & 6.4137 & 0.9892 & 2.2305 & 6.3816 \\
                                    & LightGBM& 1.0452 & 2.2764 & 6.3967 & 1.0843 & 2.2427 & \textbf{6.3814} & 0.9892 & 2.2243 & \textbf{6.3704} \\

\midrule
\rowcolor{gray!20} \textbf{MSE} & Ridge   & \textbf{1.038e-05} & \textbf{0.0008} & \textbf{0.0055} & \textbf{9.833e-06} & \textbf{0.0008} & \textbf{0.0054} & \textbf{1.023e-05} & \textbf{0.0009} & \textbf{0.0054} \\
                                & XGBoost & 1.485e-05 & 0.0010 & 0.0092 & 1.311e-05 & 0.0010 & 0.0082 & 1.232e-05 & 0.0010 & 0.0066 \\
                                & LightGBM& 1.439e-05 & 0.0009 & 0.0066 & 1.286e-05 & 0.0010 & 0.0066 & 1.255e-05 & 0.0010 & 0.0073 \\
\midrule
\rowcolor{gray!20} \textbf{MAPE (\%)} & Ridge & \textbf{1.5618} & \textbf{6.8582} & \textbf{35.7171} & \textbf{1.5501} & \textbf{6.7839} & \textbf{35.4291} & 1.5488 & \textbf{6.7871} & \textbf{35.4141} \\
                                    & XGBoost & 1.6860 & 13.1145 & 253.528 & 1.6384 & 11.9628 & 211.3958 & 1.5417 & 11.9655 & 215.0616 \\
                                    & LightGBM& 1.6836 & 11.5732 & 196.498 & 1.6439 & 11.2803 & 186.2934 & \textbf{1.5326} & 10.5628 & 160.4748 \\
\bottomrule
\end{tabular} \label{Table2}
\end{adjustbox}
\end{table}

However, for Ridge regression, instance-based clustering led to a slight decline in performance. Being a feature-transforming model, Ridge relies heavily on capturing broader, global trends, and clustering based on instance-specific characteristics diluted its ability to focus on these global patterns. As a result, its error metrics were slightly worse compared to model-based clustering, where the global scope was preserved. This indicates that while instance-based clustering improves performance for target-transforming models, it introduces challenges for feature-transforming models like Ridge, which require a more generalized dataset to maintain consistent performance.

Weighted instance-based clustering provided a balanced approach, improving the performance of both XGBoost and LightGBM even further. This method allowed the models to prioritize more relevant or informative data points within each cluster by assigning weights to instances based on their importance. This resulted in lower nMAE and MSE for both XGBoost and LightGBM. The ability to weigh instances helped these models refine their predictions by focusing on the most critical segments of the data, leading to better generalization across diverse patterns and anomalies. LightGBM, in particular, benefited from this method as the weighted approach helped mitigate the performance drop seen with model-based clustering, restoring and even improving its forecasting accuracy.

For Ridge regression, weighted instance-based clustering provided a compromise between the purely global and purely local approaches. While not as strong as model-based clustering in maintaining globality, the weighted approach allowed Ridge to balance local patterns with broader trends. Its performance in weighted instance-based clustering was more stable than in instance-based clustering, though still not as strong as in model-based clustering. The weighted method helped Ridge mitigate some of the overfitting risks associated with instance-based clustering by retaining more global relevance in the data, making it a more viable option for feature-transforming models compared to unweighted instance-based clustering.

In comparing these clustering strategies, it becomes clear that the choice of clustering method has a profound impact on the performance of global models. Model-based clustering remains the most effective approach for feature-transforming models like Ridge, which thrive on capturing broad trends. Conversely, target-transforming models like XGBoost and LightGBM perform best with weighted instance-based clustering, where they can balance local adaptability with the generalization benefits of global data. This suggests that for heterogeneous datasets, clustering strategies should be carefully selected based on the underlying model architecture to maximize forecasting accuracy and robustness.

While the proposed weighted instance-based TSC demonstrates improved forecasting performance across heterogeneous load profiles, its robustness under adversarial conditions such as data attacks remains unexplored. Evaluating how clustering-based forecasting models behave when exposed to corrupted or adversarial input (e.g., missing data, injected noise, or targeted manipulation) is an important direction for future work, particularly in critical infrastructure domains like power systems.

\subsection{Zero-shot Regional and System Load Forecasting}

\cref{Table3} provides valuable insights into the performance of GFMs applied to both regional and system-level load forecasting. These models were trained globally on area-level data, and their performance in zero-shot forecasting for different regions and the system is highlighted. By addressing the coherency of these zero-shot forecasts, this approach offers the potential to develop a unified framework for hierarchical load forecasting across all levels of transmission networks, from regional areas to the entire system. Coherency refers to the consistency between forecasts at different hierarchical levels, from the PoD level, through area and regional levels, all the way up to system-wide forecasts. For example, the sum of load forecasts for various regions should logically equal the total system load forecast. Without this coherence, discrepancies can arise, leading to inaccurate or conflicting predictions at different levels, which may misguide operational decisions \cite{ref48}.

Across all regions and the system level, the models demonstrate strong predictive power, with consistently low error values. Ridge slightly outperforms XGBoost and LightGBM, particularly in Central, Northeast, and South regional and system levels, highlighting its strong generalization across regional and system-wide forecasts. All accuracy metrics confirm that Ridge offers more accurate forecasts than XGBoost and LightGBM, especially in regions like the South, where Ridge achieves the lowest error rates. This indicates that linear regression models are more effective at managing variability in the dataset and minimizing errors at both the regional and system levels. Moreover, FB values are near zero across all models, indicating minimal systematic over- or under-estimation in the predictions. This lack of bias across all models ensures that the forecasts remain reliable and free from significant skew.

\begin{table}[t]
    \centering
    \caption{Zero-shot performance metrics for GFMs across different hierarchical levels. The nMAE is calculated by dividing the MAE by the maximum value of the normalized ground truth for each area.}
    \begin{adjustbox}{max width=\textwidth}
    \begin{tabular}{llccccccc}
    \toprule
    \textbf{Metric} & \textbf{Model} & \textbf{Calgary} & \textbf{Central} & \textbf{Edmonton} & \textbf{Northeast} & \textbf{Northwest} & \textbf{South} & \textbf{System} \\
    \midrule
    \rowcolor{gray!20} \textbf{FB}  & Ridge     & \textbf{0.0000} & 0.0005 & \textbf{0.0001} & 0.0007 & \textbf{0.0004} & \textbf{0.0003} & 0.0004 \\
                                    & XGBoost   & -0.0002 & \textbf{0.0000} & -0.0002 & \textbf{-0.0001} & 0.0014 & 0.0004 & \textbf{0.0000} \\
                                    & LightGBM  & -0.0002 & \textbf{0.0000} & -0.0002 & \textbf{-0.0002} & 0.0014 & 0.0004 & \textbf{0.0000} \\
    \midrule
    \rowcolor{gray!20} \textbf{nMAE (\%)} & Ridge    & 1.2814 & \textbf{1.6327} & 1.6327 & 2.0342 & 3.9712 & \textbf{1.8416} & \textbf{1.3208} \\
         & XGBoost   & \textbf{1.0418} & 1.7329 & \textbf{1.4726} & \textbf{2.1725} & \textbf{3.9462} & 1.9603 & 1.3827 \\
         & LightGBM  & 1.0452 & 1.7324 & 1.4728 & 2.1743 & 3.9532 & 1.9718 & 1.3924 \\

    \midrule
    \rowcolor{gray!20} \textbf{MSE} & Ridge    & 0.0003 & \textbf{0.0001} & \textbf{0.0003} & \textbf{0.00004} &  \textbf{0.0011} & \textbf{0.0003} & \textbf{0.0001} \\
                                    & XGBoost  & \textbf{0.0002} & \textbf{0.0001} & \textbf{0.0003} & 0.00005 & \textbf{0.0011} & \textbf{0.0003} & \textbf{0.0001} \\
                                    & LightGBM & \textbf{0.0002} & \textbf{0.0001} & \textbf{0.0003} & 0.00005 & \textbf{0.0011} & \textbf{0.0003} & \textbf{0.0001} \\
    \midrule
\rowcolor{gray!20} \textbf{MAPE (\%)} & Ridge  & 3.5217 & \textbf{3.3926} & 4.1576 & \textbf{3.7529} & \textbf{9.0018} & \textbf{5.6282} & 3.3711 \\
                                    & XGBoost  & \textbf{2.7843} & 3.6235 & 3.7419 & 3.9890 & 9.0432 & 6.0764 & \textbf{3.5236} \\
                                    & LightGBM & 2.7849 & 3.6237 & \textbf{3.7287} & 3.9892 & 9.0512 & 6.0865 & 3.5240 \\
\bottomrule

    \end{tabular}
    \end{adjustbox}
    \label{Table3}
\end{table}

\subsection{Peak Load Forecasting}

In this section, we investigate the performance of local, global, and cluster-wise global load forecasting models in predicting peak load events. Peak load forecasting is critical for grid management, as it directly impacts resource allocation, demand response strategies, and infrastructure planning decisions \cite{ref49}. Accurately predicting peak load is challenging due to the sporadic and extreme nature of these events, which often result from various factors such as weather conditions, seasonal demand fluctuations, and socio-economic activities \cite{ref50}. By comparing the performance of localized models, which capture area-specific patterns, with global models that generalize across regions, and cluster-wise global models that account for data heterogeneity by grouping similar regions, we aim to evaluate the efficacy of each approach in handling the complexities associated with peak load forecasting.

\begin{figure}[!t]
  \centering
  \includegraphics[width=0.84\textwidth, trim={0.15cm 0.25cm 0cm 0cm},clip]{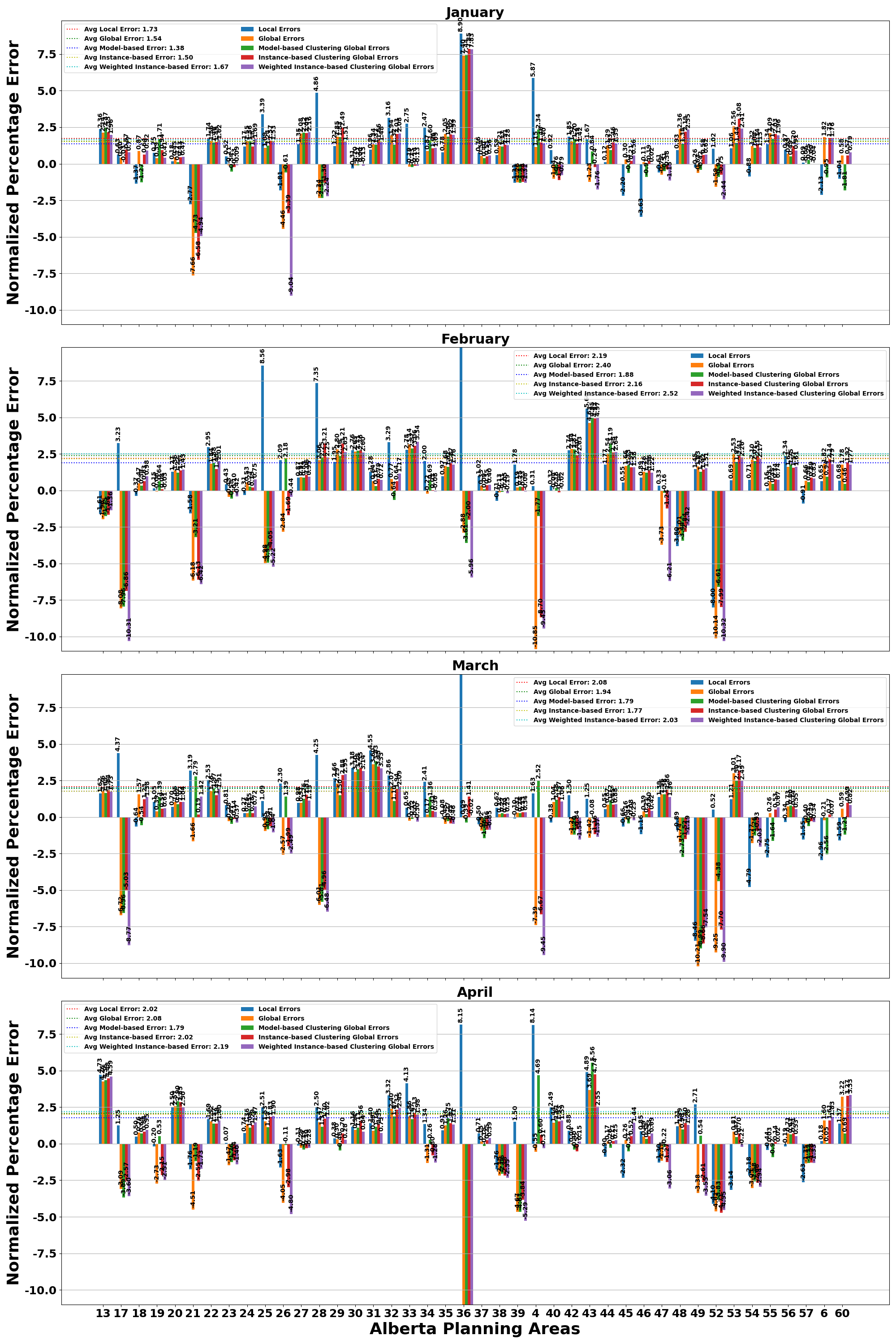}
  \caption{Monthly peak load forecasting error (Jan-Apr) using the Ridge model. The error is reported as a percentage of the actual peak load.}
  \label{fig:Fig12}
\end{figure}

\begin{figure}[!t]
  \centering
  \includegraphics[width=0.84\textwidth,trim={0.15cm 0.25cm 0cm 0cm},clip]{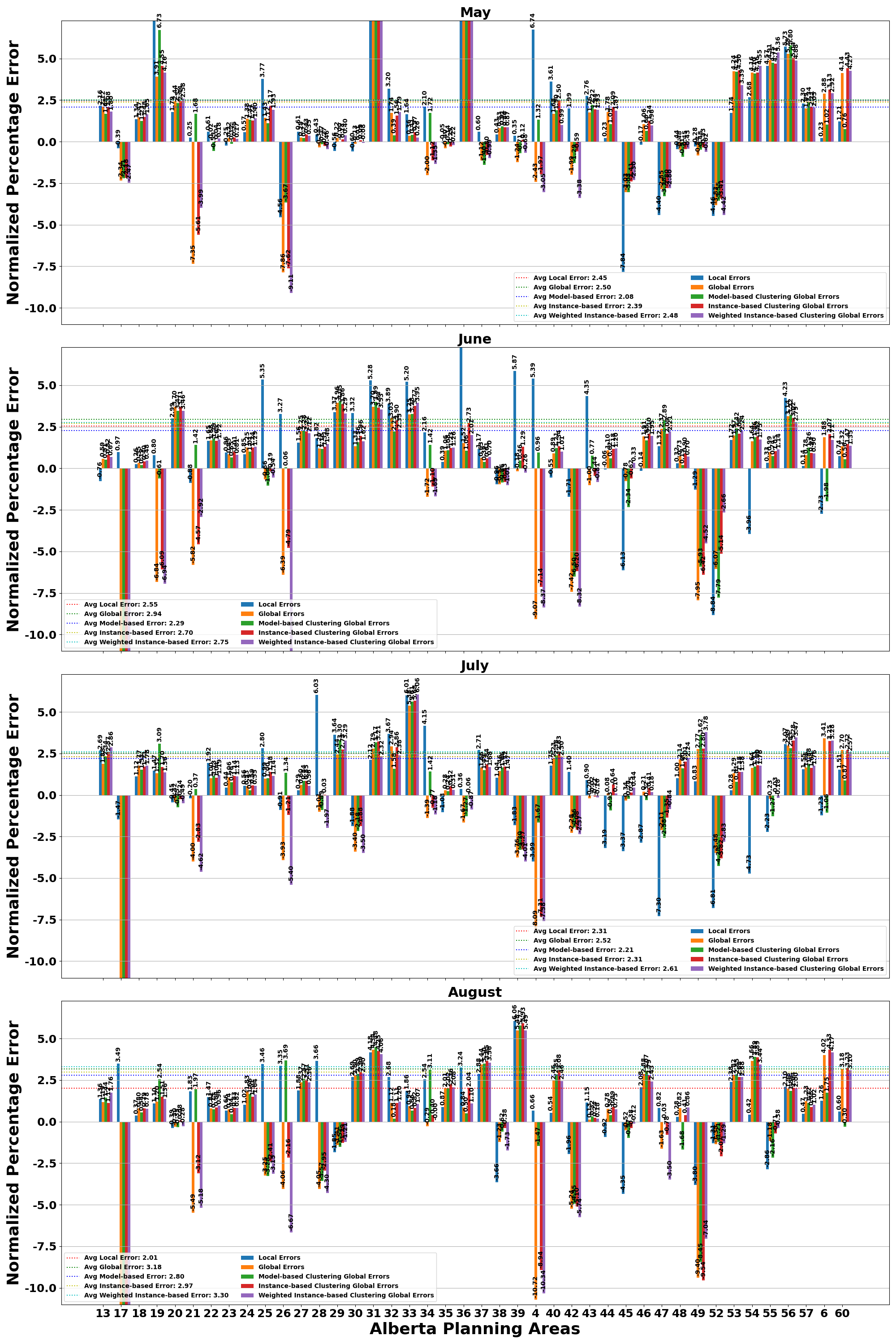}
  \caption{Monthly peak load forecasting (May-Aug) using the Ridge model. The error is reported as a percentage of the actual peak load.}
  \label{fig:Fig13}
\end{figure}

\begin{figure}[!t]
  \centering
  \includegraphics[width=0.84\textwidth,trim={0.15cm 0.25cm 0cm 0cm},clip]{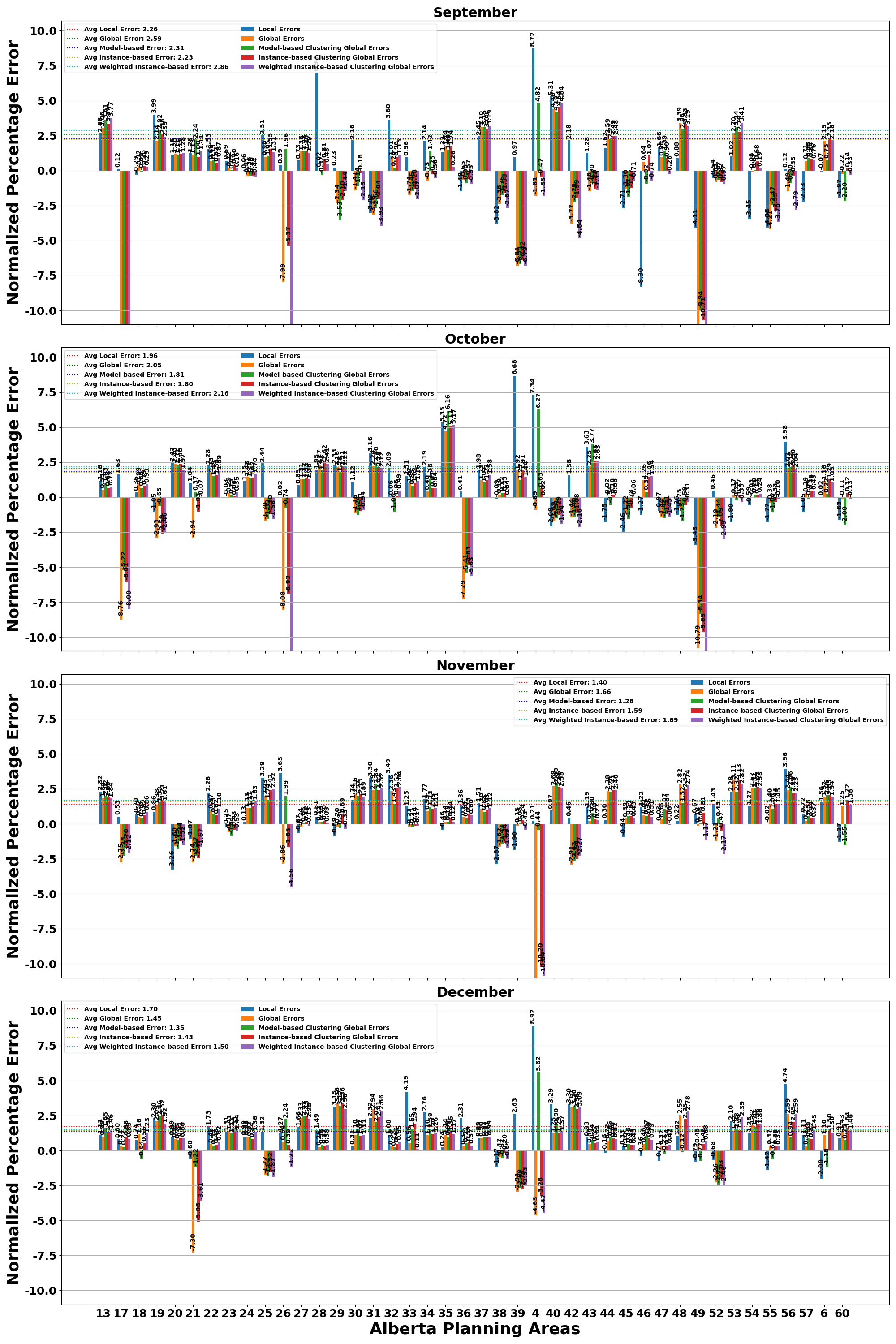}
  \caption{Monthly peak load forecasting (Sep-Dec) using the Ridge model. The error is reported as a percentage of the actual peak load.}
  \label{fig:Fig14}
\end{figure}

\begin{figure}[!t]
  \centering
  \includegraphics[width=0.84\textwidth,trim={0.15cm 0.25cm 0cm 0cm},clip]{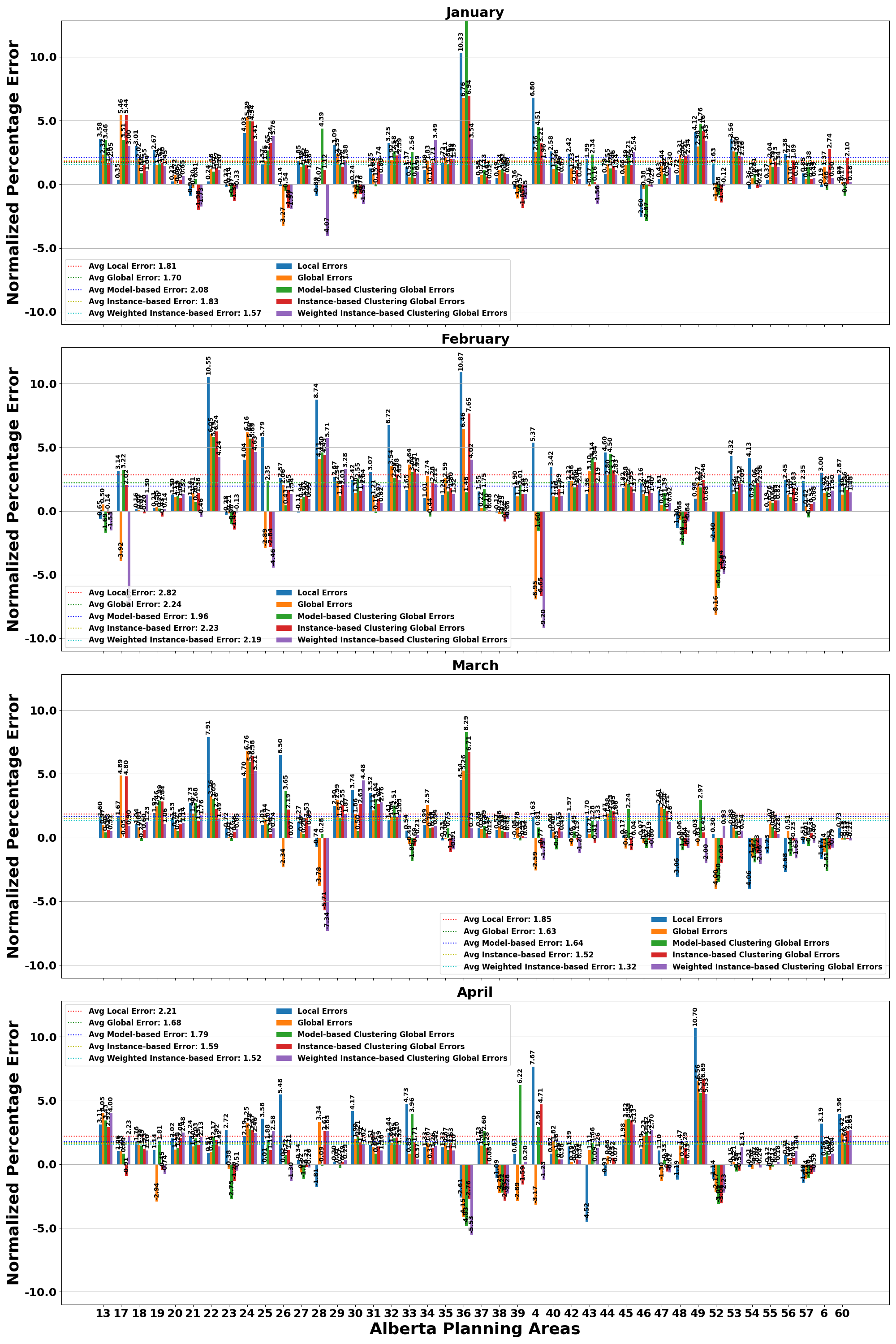}
  \caption{Monthly peak load forecasting (Jan-Apr) using the LightGBM model. The error is reported as a percentage of the actual peak load.}
  \label{fig:Fig15}
\end{figure}

\begin{figure}[!t]
  \centering
  \includegraphics[width=0.84\textwidth,trim={0.15cm 0.25cm 0cm 0cm},clip]{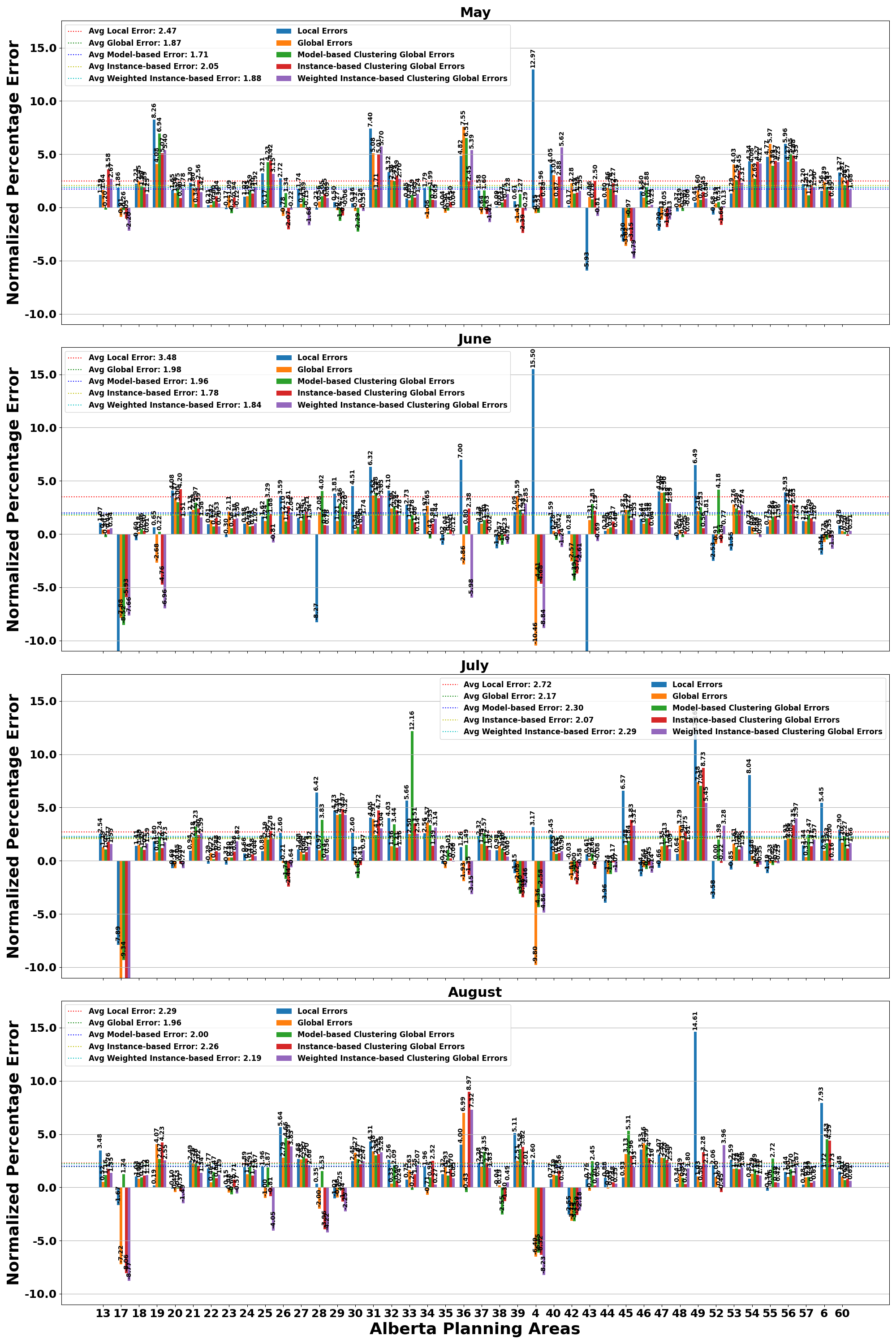}
  \caption{Monthly peak load forecasting (May-Aug) using the LightGBM model. The error is reported as a percentage of the actual peak load.}
  \label{fig:Fig16}
\end{figure}

\begin{figure}[!t]
  \centering
  \includegraphics[width=0.84\textwidth,trim={0.15cm 0.25cm 0cm 0cm},clip]{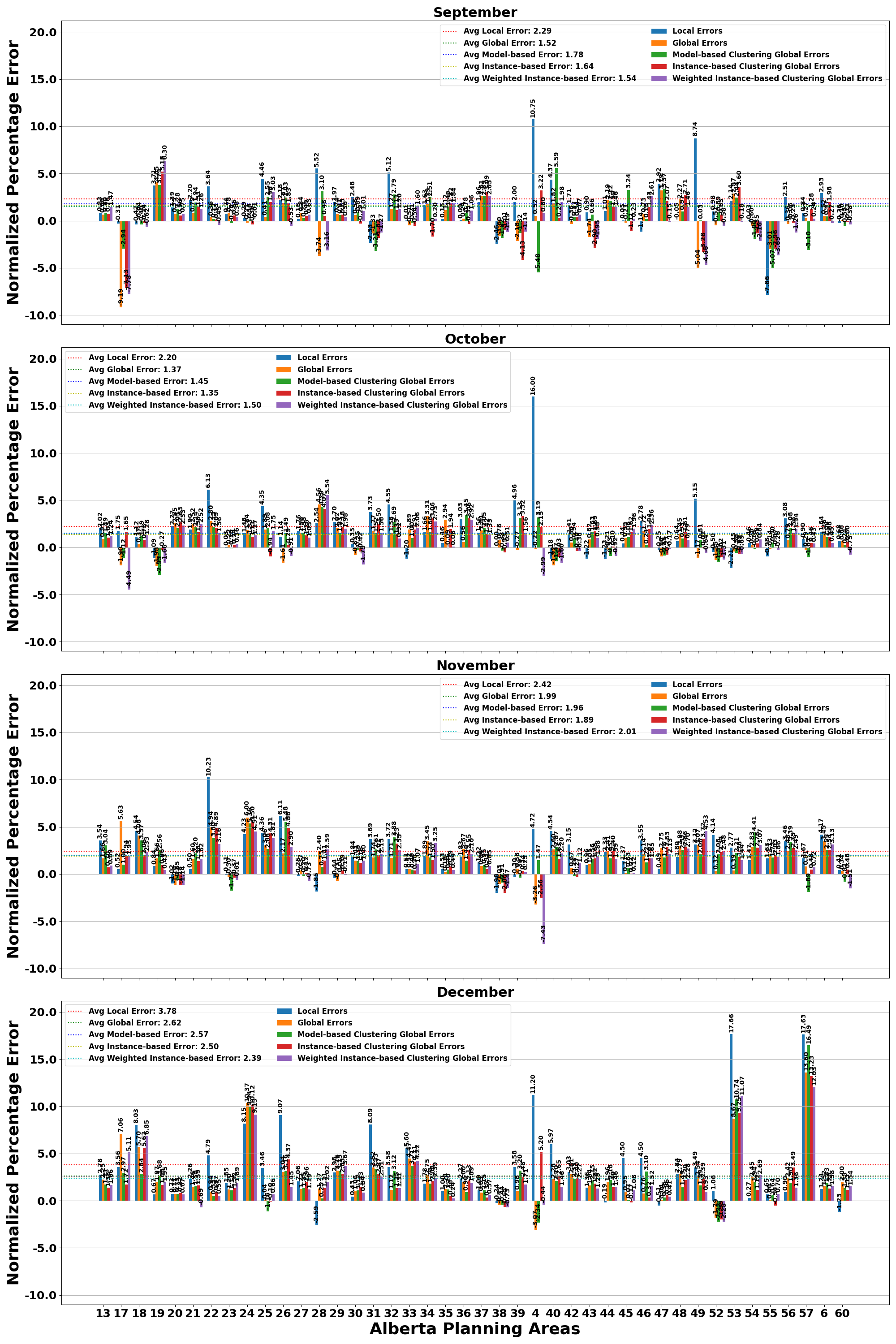}
  \caption{Monthly peak load forecasting (Sep-Dec) using the LightGBM model. The error is reported as a percentage of the actual peak load.}
  \label{fig:Fig17}
\end{figure}

\cref{fig:Fig12,fig:Fig13,fig:Fig14,fig:Fig15,fig:Fig16,fig:Fig17} illustrate the comparative performance of peak load forecasting using the Ridge and LightGBM models, respectively, across different months of the year, employing various modeling strategies: local, global, and clustering-based techniques. Each sub-figure represents the error distribution for a specific month, with the vertical axis indicating the forecasting error for different areas within the region, and the horizontal axis representing the respective area indices.

Across all months, the global model consistently demonstrates superior performance compared to the local models. This result suggests that having access to a larger dataset allows the global model to capture the underlying patterns of peak load demand more effectively, leading to improved forecasting accuracy. The global model’s ability to generalize across diverse areas with varying load characteristics highlights the benefit of leveraging a comprehensive dataset that encompasses multiple regions, reducing the limitations faced by local models that rely on less data and may struggle with variability.

\begin{figure*}[!t]
  \centering
  \includegraphics[width=0.95\textwidth,trim={0.15cm 0.25cm 0cm 0cm},clip]{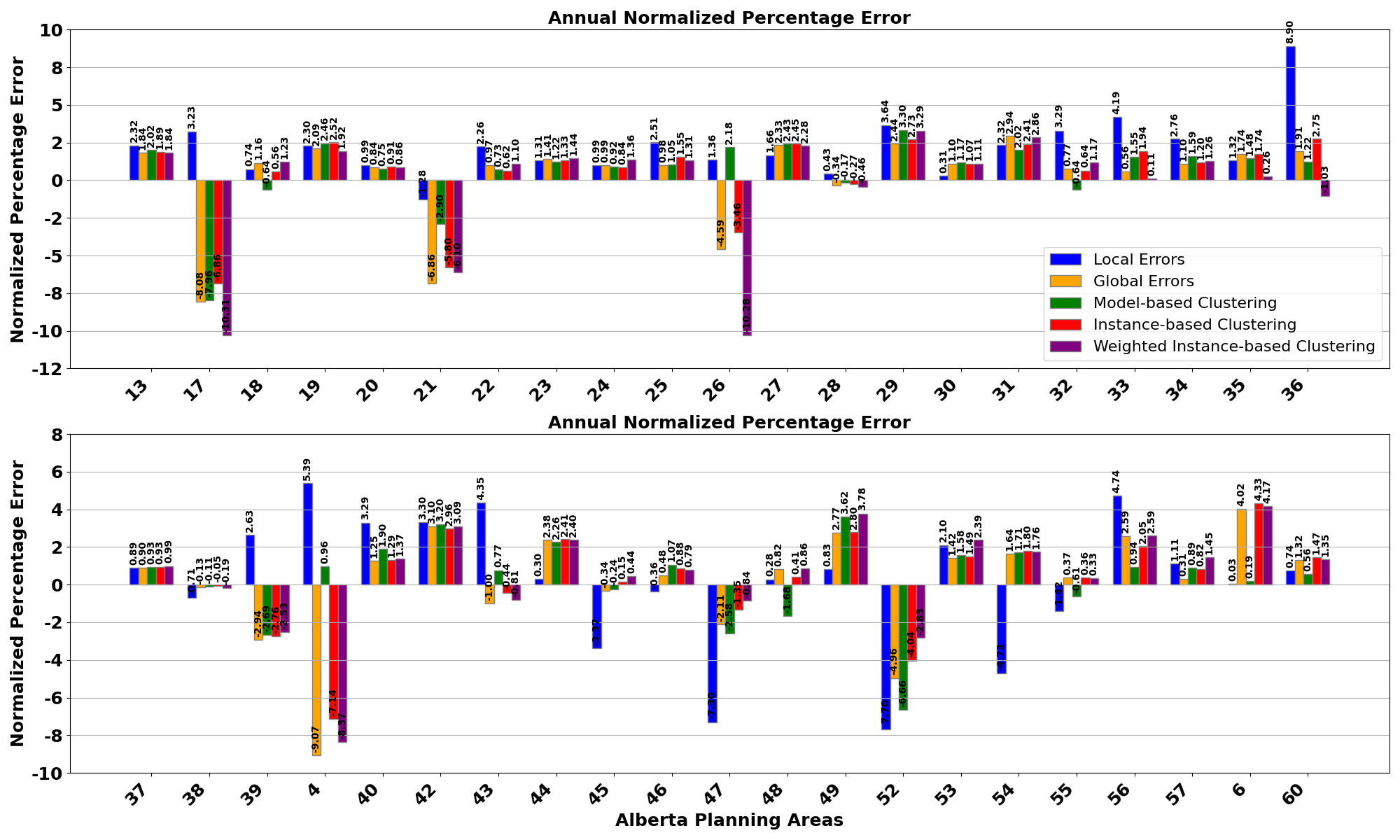}
  \caption{Annual peak load forecasting using the Ridge model. The error is reported as a percentage of the actual peak load. On average, the Local model had an error of 2.47\%, the Global model 2.12\%, the Model-based Clustering Global model 1.75\%, the Instance-based Clustering Global model 1.99\%, and the Weighted Instance-based Clustering Global model 2.26\%.}
  \label{fig:Fig18}
\end{figure*}

\begin{figure*}[!t]
  \centering
  \includegraphics[width=0.95\textwidth,trim={0.15cm 0.25cm 0cm 0cm},clip]{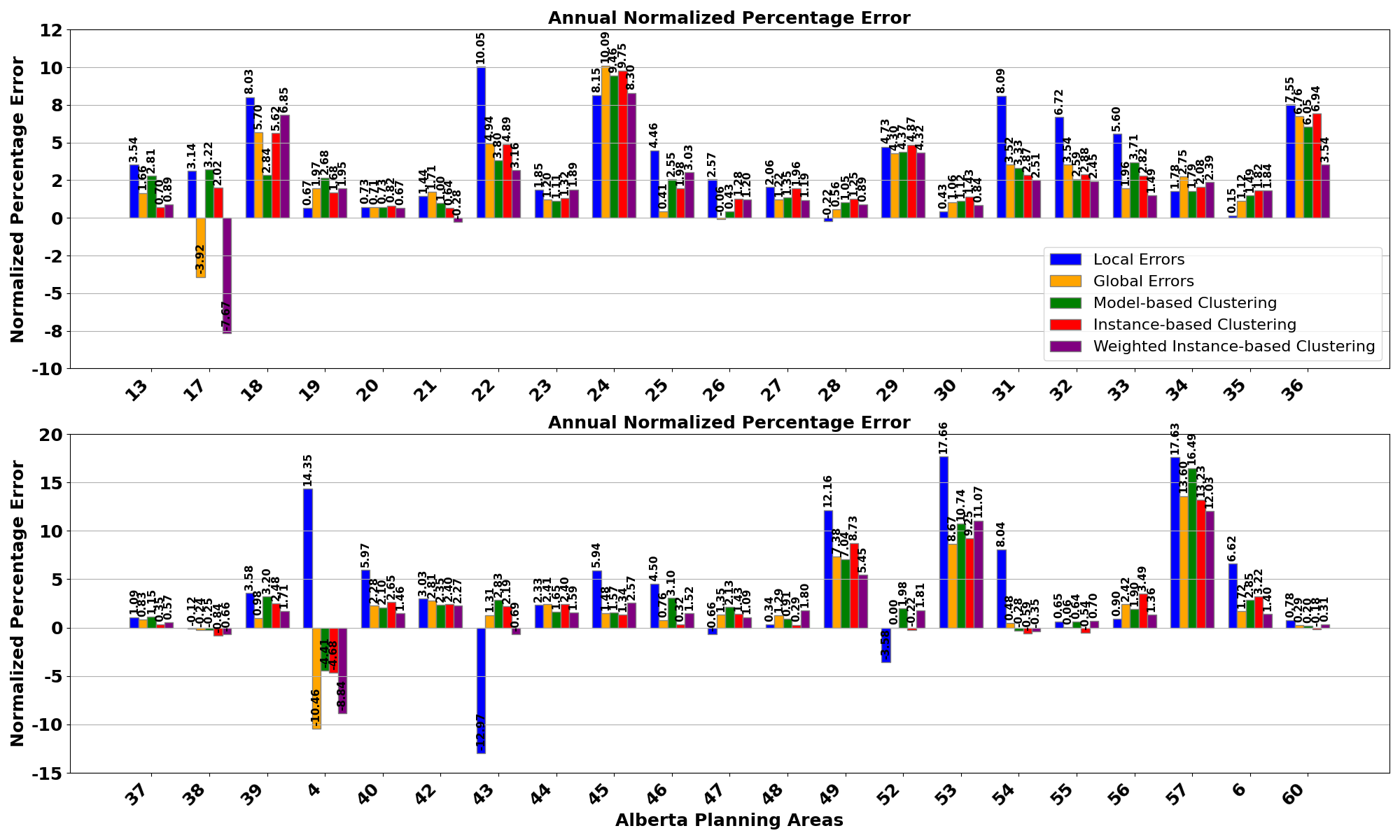}
  \caption{Annual peak load forecasting using the LightGBM model. The error is reported as a percentage of the actual peak load. On average, the Local model had an error of 4.88\%, the Global model 2.86\%, the Model-based Clustering Global model 2.98\%, the Instance-based Clustering Global model 2.87\%, and the Weighted Instance-based Clustering Global model 2.78\%.}
  \label{fig:Fig19}
\end{figure*}

\cref{fig:Fig18,fig:Fig19} illustrate the results of annual peak load forecasting using different modeling strategies, including local, global, and clustering-based techniques, applied to various Alberta planning areas. The overall results demonstrate that the global model consistently outperforms the local models in terms of average error across all areas. This suggests that utilizing a larger dataset encompassing diverse regions enables the global model to gain a more comprehensive view, thereby improving its ability to generalize and produce more accurate peak load forecasts. The local models, on the other hand, show higher errors, particularly in areas with more complex and volatile load patterns. This suggests that local models may be limited by the smaller amount of data available for training, making it challenging to effectively capture annual variations.

An interesting observation is the consistent performance of the model-based clustering global Ridge model, which outperforms not only its local and global Ridge but also other clustering-based global Ridge models. This trend is similarly observed with the weighted instance-based clustering approach in LightGBM, which outperforms not only its local and global LightGBM but also other clustering-based global Ridge models. The superior performance of these clustering techniques suggests that appropriately grouping areas with similar characteristics can significantly enhance forecasting accuracy, particularly in complex and heterogeneous data scenarios.

Overall, the results emphasize the strength of the global model for peak load forecasting. At the same time, clustering-based methods provide a nuanced approach to capturing local variations, enabling the global model to compete effectively with localized models. These findings underscore the importance of selecting the appropriate modeling strategy based on the data characteristics and the specific forecasting requirements.

\section{Conclusion} \label{sec:Conclusions}

In this paper, we explored the limitations of traditional LFMs in the context of transmission networks and highlighted the advantages of GFMs for scalable load prediction. Our investigation into GFMs reveals their capacity to improve prediction accuracy and generalizability by leveraging cross-learning across multiple time series, relieving challenges such as data drift, and overcoming the scalability issues inherent in LFMs.

Moreover, GFMs hold significant potential for application in hierarchical forecasting due to their ability to benefit from cross-learning across different time series. However, for GFMs to fully realize this potential, they must learn to maintain coherence between their forecasts across various hierarchical levels, ensuring consistency across time series.

However, like LFMs, GFMs trained for single-step-ahead forecasting often fail to capture long-term behaviors, such as seasonality, and tend to mimic naïve forecasts. GFMs trained on longer prediction horizons are expected to resolve this issue, as they are forced to learn more comprehensive patterns in the data. The same holds true for their probabilistic counterparts, which benefit from larger datasets, resulting in greater certainty in their predictions. Addressing points, particularly in the context of longer-horizon forecasting and improving the probabilistic versions of global models, will be a focus of our future work.

This study also emphasizes the complexities associated with GFMs, particularly concerning data heterogeneity and the assumption that input time series are related. We also analyzed different modeling techniques, demonstrating how globalization, data heterogeneity, and data drift affect each differently. To tackle data heterogeneity, we proposed model-based TSC for feature-transformers and weighted instance-based TSC for target-transformers. This aimed to balance the trade-off between globality and locality by ensuring that the GFMs are tailored to capture the unique characteristics of each cluster, thereby enhancing GFM's performance.

In addition to proposing clustering-based solutions, we analyzed regional disparities in forecasting performance to understand better where GFMs face challenges. We found that areas such as Area 4 (city of Medicine Hat), Areas 21 and 26, and Area 49 consistently exhibited degraded performance across models. Area 4 has undergone substantial behind-the-meter load growth due to increased industrial and commercial activities like such as cryptocurrency mining, leading to structural shifts that reduce generalizability. Areas 21 and 26 are influenced by demand response behaviors, which introduce behavioral noise and complicate forecasts. Moreover, Area 49 experienced a marked increase in load volatility following the COVID-19 pandemic, which further hinders forecasting accuracy. These findings underscore the importance of designing heterogeneity-aware forecasting strategies that explicitly account for spatial, behavioral, and temporal variability across regions.

While the proposed weighted instance-based TSC demonstrates improved forecasting performance across heterogeneous load profiles, its robustness under adversarial conditions such as data attacks remains unexplored. Evaluating how clustering-based forecasting models behave when exposed to corrupted or adversarial input (e.g., missing data, injected noise, or targeted manipulation) is an important direction for future work, particularly in critical infrastructure domains like power systems.

Furthermore, we extended our analysis to the ISO-NE dataset (Hong et al., 2019), which represents a relatively homogeneous system across its zones. As detailed in the Appendix, our results show that in the absence of significant heterogeneity, the global forecasting model alone performs better than both local models. This confirms that clustering is not always necessary: in the homogeneous settings, a simple global model can leverage shared structure efficiently without added complexity. This finding reinforces the adaptability of global models and underscores the importance of quantifying spatial heterogeneity before selecting model architectures.

\appendix
\section{Globalization in Homogeneous Time Series}

Findings from Shao et al. \cite{ref455} highlight that temporal characteristics, such as stability, seasonality, and distributional drift, and spatial dependencies, particularly those quantified through spatial indistinguishability metrics, play a critical role in shaping the effectiveness of forecasting model architectures. Specifically, in the presence of strong heterogeneity TSC offers significant advantages. In contrast, when data across time series are relatively homogeneous, global models tend to perform equally well or even outperform more complex or partitioned approaches due to their simplicity and generalization ability.

To investigate the latter scenario, we conduct supplementary experiments using the publicly available dataset from the 2017 Global Energy Forecasting Competition (GEFCom2017). It comprises over 14 years (March 2003 to April 2017) of hourly electricity load and weather data from the ISO New England control area. The dataset covers ten geographical zones, eight bottom-level and two aggregated, and includes hourly measurements of electricity load, dry-bulb temperature, and dew point temperature for each zone.

\begin{figure}[!t]
  \centering
  \includegraphics[width=0.95\textwidth,trim={0.15cm 0.25cm 0.15cm 1.5cm},clip]{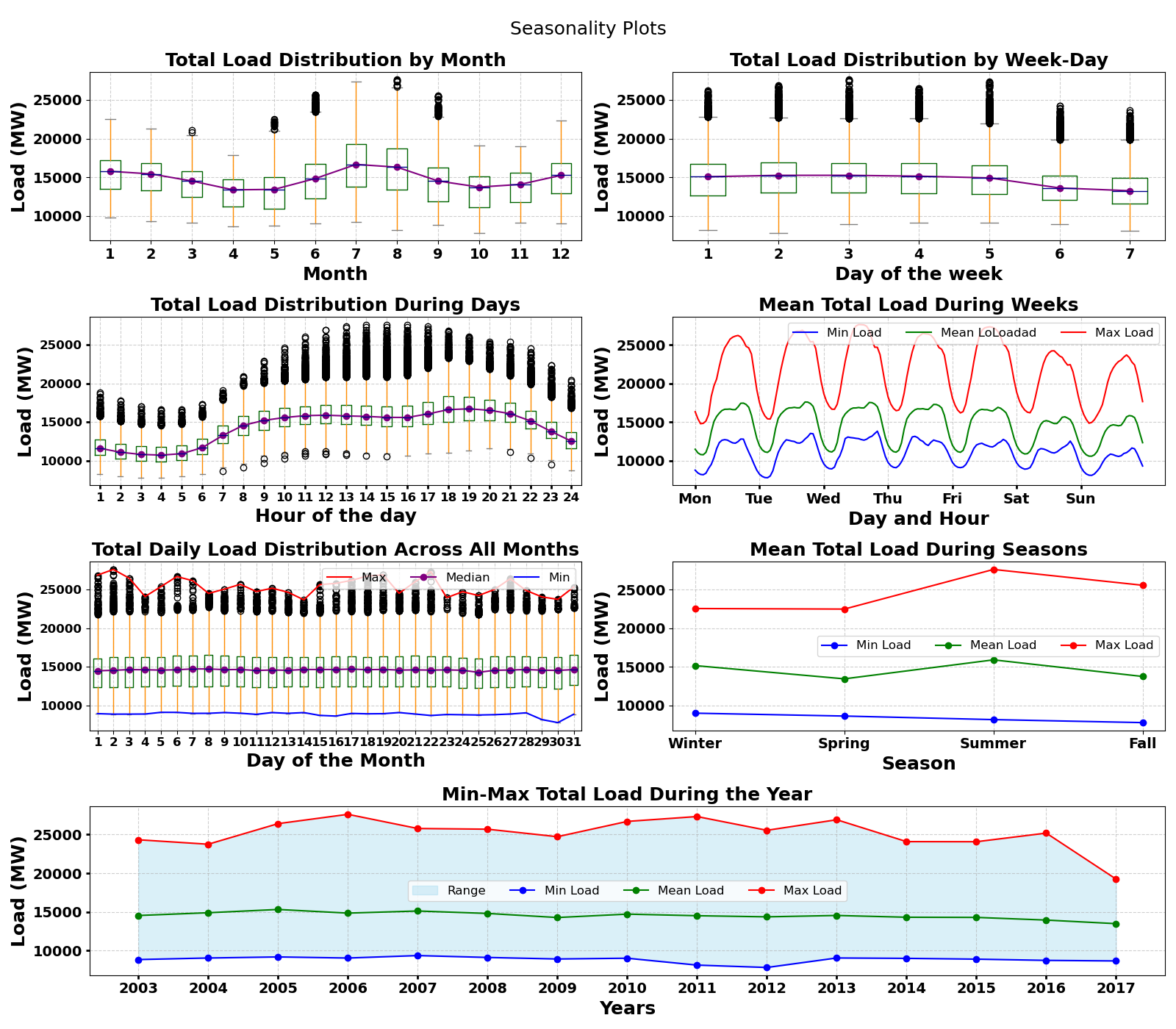}
  \caption{Total load profile of ISO New England control area.}
  \label{fig:Fig20}
\end{figure}

\cref{fig:Fig20} presents seasonality plots aggregated across all zones. Load displays strong and consistent seasonal, weekly, and daily patterns, peaking in summer, during weekdays, and in daytime hours. These temporal patterns remain stable over the 14-year period, indicating high temporal regularity across the system. The overlapping histograms for training (blue) and test (red) sets also indicate that both sets share similar distributions for all zones, suggesting no data drift.

\begin{figure}[!t]
  \centering
  \includegraphics[width=1\textwidth,trim={0.15cm 0.25cm 0cm 0cm},clip]{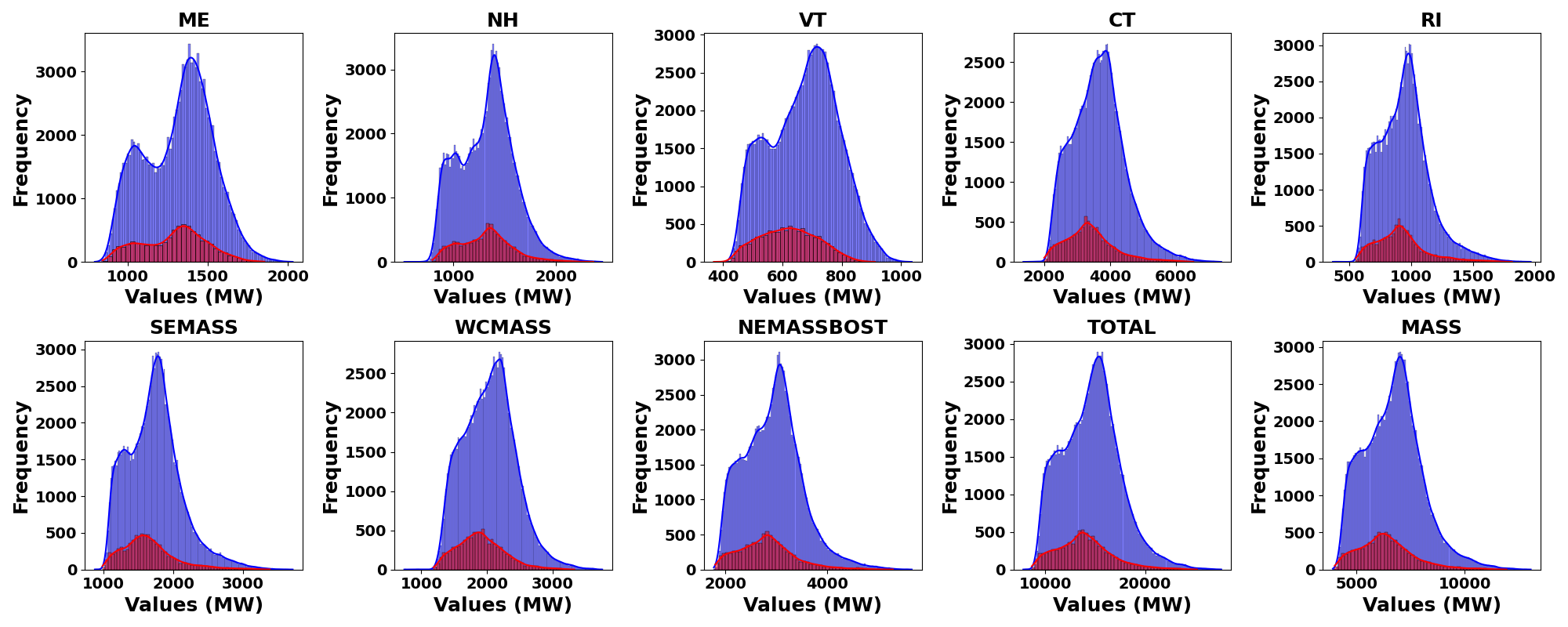}
  \caption{Load distribution across all ten GEFCom2017 zones.}
  \label{fig:Fig21}
\end{figure}

\cref{fig:Fig21} shows load histograms for all ten zones. Despite differences in absolute load levels, the shape, skewness, and modality of the distributions are highly similar across zones. This statistical resemblance suggests that all zones follow comparable underlying load dynamics, supporting the assumption of spatial homogeneity.

\begin{figure}[!t]
  \centering
  \includegraphics[width=1\textwidth,trim={0.15cm 0.25cm 0cm 0cm},clip]{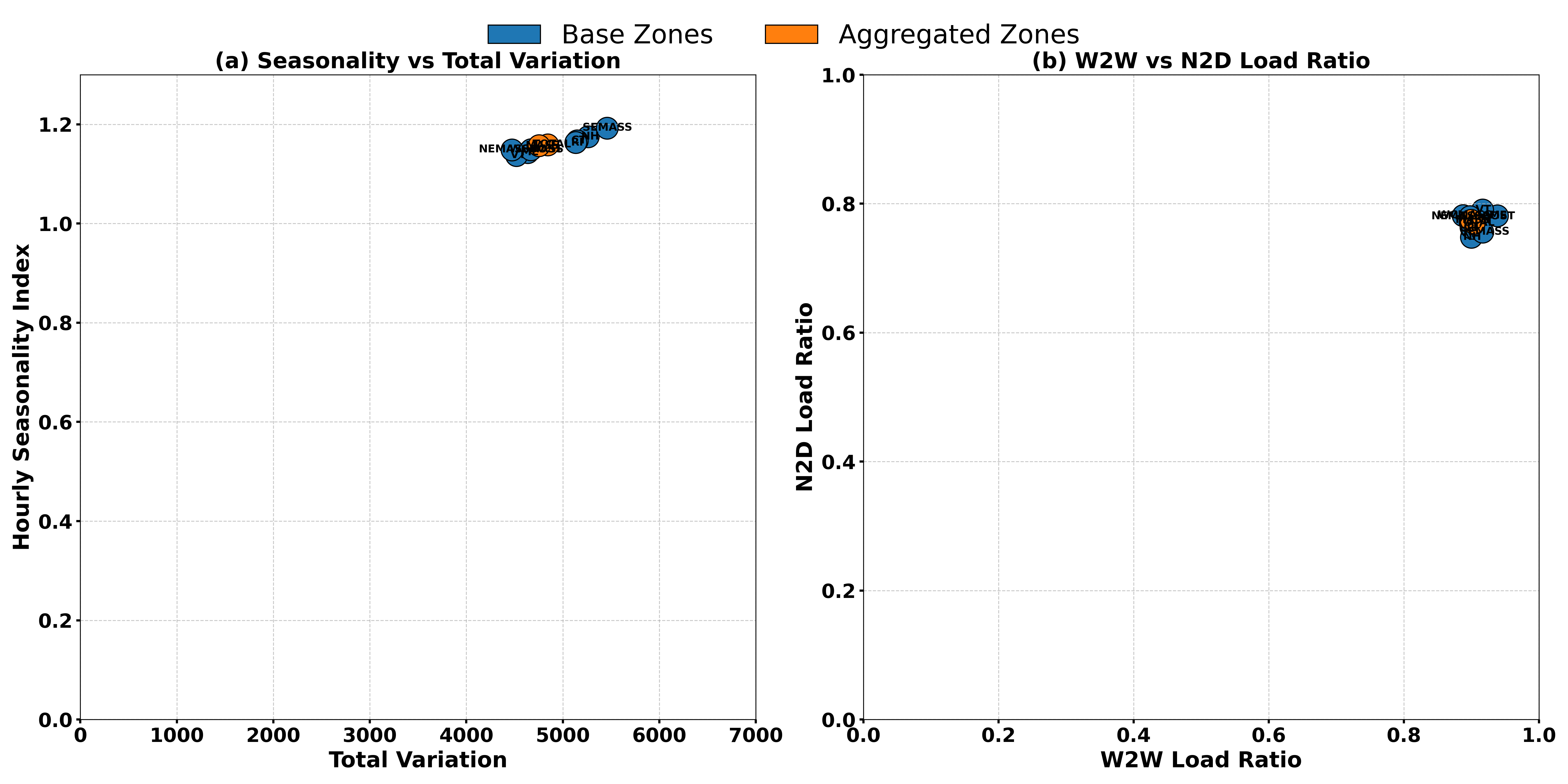}
  \caption{Consumption pattern characteristics across zones.}
  \label{fig:Fig22}
\end{figure}

\cref{fig:Fig22} reinforces this finding by comparing all zones using key seasonality metrics. In panel (a), all zones cluster tightly in the space defined by Total Variation and Hourly Seasonality Index. In panel (b), zones also align closely in terms of Weekend-to-Weekday and Night-to-Day Load Ratios. This tight clustering across multiple metrics confirms that the zones do not exhibit distinct consumption patterns that would necessitate clustering-based segmentation.

\cref{Table4} provides a comparative evaluation of LFMs and GFMs across three representative model types: Ridge Regression, XGBoost, and LightGBM. Across all metrics, GFMs consistently match or outperform LFMs. Notably, Ridge GFMs exhibit lower mean nMAE (0.77\% vs. 0.82\%) and MAPE (2.38\% vs. 2.59\%) compared to their local counterparts. Similar improvements are observed for the tree-based models, suggesting that the use of a single global model not only reduces modeling complexity but also enhances performance in homogeneous settings like GEFCom2017. These results support our earlier visual analysis and reinforce the idea that clustering or local specialization is not always necessary when the constituent time series exhibit homogeneity.

\begin{table}[t]
\centering
\caption{Normalized Performance Metrics for LFMs, and GFMs. The nMAE is calculated by dividing the MAE by the maximum value of the normalized ground truth for each area.}
\begin{adjustbox}{max width=\textwidth}
\begin{tabular}{llccccccccccc}
\toprule
\textbf{Metric} & \textbf{Model} & \multicolumn{3}{c}{\textbf{LFMs}} & \multicolumn{3}{c}{\textbf{GFMs}} \\
\cmidrule(lr){3-5} \cmidrule(lr){6-8}
 & & \textbf{Min} & \textbf{Mean} & \textbf{Max} & \textbf{Min} & \textbf{Mean} & \textbf{Max} \\
\midrule
\rowcolor{gray!20} \textbf{FB} & Ridge & 0.0001 & 0.0002 & 0.0004 & -0.0007 & 0.0002 & 0.0007 \\
                             & XGBoost & -0.0015 & -0.0003 & 0.0001 & 0.0000 & 0.0001 & 0.0004 \\
                            & LightGBM & -0.0015 & -0.0003 & 0.0002 & 0.0000 & 0.0002 & 0.0004 \\
\midrule
\rowcolor{gray!20} \textbf{nMAE (\%)} & Ridge & 0.63   & 0.82   & 1.49   & 0.60   & 0.77   & 1.34 \\
                                  & XGBoost   & 0.90   & 1.19   & 2.03   & 0.83   & 1.08   & 1.77 \\
                                  & LightGBM  & 0.89   & 1.19   & 2.06   & 0.83   & 1.08   & 1.78 \\
\midrule
\rowcolor{gray!20} \textbf{MSE} & Ridge & 0.0001 & 0.0001 & 0.0003 & 0.0001 & 0.0001 & 0.0002 \\
                              & XGBoost & 0.0001 & 0.0002 & 0.0005 & 0.0001 & 0.0002 & 0.0004 \\
                             & LightGBM & 0.0001 & 0.0002 & 0.0005 & 0.0001 & 0.0002 & 0.0004 \\
\midrule
\rowcolor{gray!20} \textbf{MAPE (\%)} & Ridge & 1.66   & 2.59   & 3.92   & 1.52   & 2.38   & 3.43 \\
                                   & XGBoost  & 2.39   & 3.97   & 7.18   & 2.03   & 3.53   & 6.56 \\
                                   & LightGBM & 2.35   & 3.96   & 7.20   & 2.05   & 3.50   & 6.44 \\
\bottomrule
\end{tabular} \label{Table4}
\end{adjustbox}
\end{table}




\end{document}